\documentclass[10pt,twocolumn,letterpaper]{article}

\usepackage[pagenumbers]{cvpr} 

\usepackage{graphicx}
\usepackage{amsmath}
\usepackage{amssymb}
\usepackage{booktabs}
\usepackage[accsupp]{axessibility}

\usepackage[pagebackref,breaklinks,colorlinks]{hyperref}

\usepackage[capitalize]{cleveref}
\crefname{section}{Sec.}{Secs.}
\Crefname{section}{Section}{Sections}
\Crefname{table}{Table}{Tables}
\crefname{table}{Tab.}{Tabs.}

\def\cvprPaperID{6919} 
\def\confName{CVPR}
\def\confYear{2023}

\usepackage{color, colortbl}
\usepackage{amsmath}
\usepackage{xcolor}
\usepackage[linesnumbered,ruled,vlined]{algorithm2e}
\usepackage{graphicx,wrapfig,lipsum}

\SetCommentSty{mycommfont}
\SetKwInput{KwInput}{Input}                
\SetKwInput{KwOutput}{Output}              

\newcommand\norm[1]{\lVert#1\rVert}

\DeclareMathOperator*{\argmin}{arg\,min}

\definecolor{Gray}{gray}{0.9}

\begin{document}

\title{Achieving a Better Stability-Plasticity Trade-off via Auxiliary Networks in Continual Learning}

\author{Sanghwan Kim \quad  Lorenzo Noci  \quad Antonio Orvieto  \quad Thomas Hofmann\\
ETH Zürich\\
Zürich, Switzerland\\
{\tt\small \{sanghwan.kim, lorenzo.noci, antonio.orvieto, thomas.hofmann\}@inf.ethz.ch}
}
\maketitle

\begin{abstract}
In contrast to the natural capabilities of humans to learn new tasks in a sequential fashion, neural networks are known to suffer from \emph{catastrophic forgetting}, where the model's performances on old tasks drop dramatically after being optimized for a new task. Since then, the continual learning (CL) community has proposed several solutions aiming to equip the neural network with the ability to learn the current task (\emph{plasticity}) while still achieving high accuracy on the previous tasks (\emph{stability}). Despite remarkable improvements, the plasticity-stability trade-off is still far from being solved and its underlying mechanism is poorly understood. In this work, we propose \emph{Auxiliary Network Continual Learning} (ANCL), a novel method that applies an additional auxiliary network which promotes plasticity to the continually learned model which mainly focuses on stability. More concretely, the proposed framework materializes in a regularizer that naturally interpolates between plasticity and stability, surpassing strong baselines on task incremental and class incremental scenarios. Through extensive analyses on ANCL solutions, we identify some essential principles beneath the stability-plasticity trade-off. The code implementation of our work is available at \url{https://github.com/kim-sanghwan/ANCL}.
\end{abstract}

\section{Introduction}\label{sec:introduction}

The continual learning (CL) model aims to learn from current data while still maintaining the information from previous training data. The naive approach of continuously fine-tuning the model on sequential tasks, however, suffers from \emph{catastrophic forgetting} \cite{mccloskey1989catastrophic, goodfellow2013empirical}. Catastrophic forgetting occurs in a gradient-based neural network because the updates made with the current task are likely to override the model weights that have been changed by the gradients from the old tasks.

Catastrophic forgetting can be understood in terms of \textit{stability-plasticity dilemma} \cite{mermillod2013stability}, one of the well-known challenges in continual learning. Specifically, the model not only has to generalize well on past data (\textit{stability}) but also learn new concepts (\textit{plasticity}). Focusing on stability will hinder the neural network from learning the new data, whereas too much plasticity will induce more forgetting of the previously learned weights. Therefore, CL model should strike a balance between stability and plasticity. 

There are various ways to define the problem of CL. Generally speaking, it can be categorized into three scenarios \cite{van2019three} : \emph{Task Incremental Learning} (TIL), \emph{Domain Incremental Learning} (DIL), and \emph{Class Incremental Learning} (CIL). In TIL, the model is informed about the task that needs to be solved; the task identity is given to the model during the training session and the test time. In DIL, the model is required to solve only one task at hands without the task identity. In CIL, the model should solve the task itself and infer the task identity. Since the model should discriminate all classes that have been seen so far, it is usually regarded as the hardest continual learning scenario. Our study performs extensive evaluations on TIL and CIL setting which will be further explained in \cref{sec:experiment}.

Recently, several papers~\cite{wang2021afec, zhang2020class, liu2021adaptive, lin2022towards} proposed the usage of an auxiliary network or an extra module that is solely trained on the current dataset, with the purpose of combining this additional structure with the previous network or module that has been continuously trained on the old datasets. For example, \emph{Active Forgetting with synaptic Expansion-Convergence} (AFEC) \cite{wang2021afec} regularizes the weights relevant to the current task through a new set of network parameters called the expanded parameters based on weight regularization methods. The expanded parameters are solely optimized on the current task and are allowed to forget the previous ones. As a result, AFEC can reduce potential negative transfer by selectively merging the old parameters with the expanded parameters. The stability-plasticity balance in AFEC is adjusted via hyperparameters which scale the regularization terms for remembering the old tasks and learning the new tasks. 

The authors of the above papers propose to mitigate the stability-plasticity dilemma by infusing plasticity through the auxiliary network or module (detailed explanation in \cref{Appendix:A}). However, a precise characterization of the interactive mechanism between the previous model and the auxiliary model is still missing in the literature. Therefore, in this paper, we first formalize the framework of CL that adopts the auxiliary network called \emph{Auxiliary Network Continual Learning} (ANCL). Given this environment, we then investigate the stability-plasticity trade-off through various analyses from both a theoretical and empirical point of view.

Our main contributions can be summarized as follows:
\begin{itemize}
    \item We propose the framework of \emph{Auxiliary Network Continual Learning} (ANCL) that can naturally incorporate the auxiliary network into a variety of CL approaches as a plug-in method (\cref{subsec:formulation_of_ANCL}).

    \item We empirically show that ANCL outperforms existing CL baselines on both CIFAR-100 \cite{krizhevsky2009learning} and Tiny ImageNet \cite{le2015tiny} (\cref{sec:experiment}).   
    \item Furthermore, we perform three analyses to investigate the stability-plasticity trade-off within ANCL (\cref{sec:trade-off_analysis}): \emph{Weight Distance}, \emph{Centered Kernel Alignment}, and \emph{Mean Accuracy Landscape}.
\end{itemize}

\section{Related Work}
\label{sec:related_work}
Continual learning approaches can be roughly categorized into weight regularization \cite{kirkpatrick2017overcoming, aljundi2018memory, chaudhry2018riemannian, wang2021afec}, knowledge distillation \cite{li2017learning, jung2016less, dhar2019learning, zhang2020class}, memory replay \cite{rebuffi2017icarl, castro2018end}, bias correction \cite{wu2019large, zhao2020maintaining, hou2019learning, douillard2020podnet}, and dynamic structure \cite{liu2021adaptive, abati2020conditional, yan2021dynamically}.

\textbf{Weight Regularization Method: } A standard way to alleviate catastrophic forgetting is to include a regularization term which binds the dynamics of each network's parameter to the corresponding parameter of the old network. For example, \textit{Elastic Weight Consolidation} (EWC) \cite{kirkpatrick2017overcoming} calculates the regularizer through the approximation of Fisher Information Matrix (FIM). \textit{Memory Aware Synapses} (MAS) \cite{aljundi2018memory} proposes the regularizer which accumulates the changes of each parameter throughout the update history. Recently, \cite{wang2021afec} suggests a biologically inspired argument to propose Active Forgetting with synaptic Expansion-Convergence (AFEC) where an additional regularization term associated with expanded parameters (or auxiliary network) is added to the loss of EWC.

\textbf{Knowledge Distillation Method: } A separate line of work adopts knowledge distillation \cite{bucilua2006model, hinton2015distilling} which was originally designed to train a more compact student network from a larger teacher network. In this way, the main network can emulate the activation or logit of the previous (or old) network while learning a new task. For instance, \textit{Learning without Forgetting} (LwF) \cite{li2017learning} proposes to learn the soft target generated by the old network while \textit{less-forgetting learning} (LFL) \cite{jung2016less} regularizes the difference between the activations of the main network and the old network. Based on LwF, \textit{Learning without Memorizing} (LwM) \cite{dhar2019learning} takes advantage of the attention of the previous network to train the current network. A recent distillation approach called \textit{Deep Model Conolidation} (DMC) \cite{zhang2020class}  proposes \textit{double distillation loss} to resolve the asymmetric property of training between old and new classes using a new network (or auxiliary network) and an unlabeled auxiliary dataset.

\textbf{Memory Replay Method: } Unlike the previous methods, replay-based methods keep a part of the previous data (or exemplars) in a memory buffer. Then, a model is trained on the current dataset and the previous exemplars to prevent the forgetting of the previous tasks. \emph{Incremental Classifier and Representation Learning} (iCaRL) \cite{rebuffi2017icarl} proposes the usage of the memory buffer derived from LwF \cite{li2017learning}. Then, iCaRL calculates the mean feature representations for each class and selects the exemplars iteratively so that the mean of the exemplars is closer to the class mean in feature space, which is called \emph{herding} sampling strategy. Another replay-based approach named \emph{End-to-End Incremental Learning} (EEIL) \cite{castro2018end} introduces an additional stage called \emph{balanced training} to fine-tune the model on a balanced dataset. The balanced dataset consists of the equal number of exemplars from each class that have been seen so far.

\textbf{Bias Correction Method: } In memory replay methods, the network is trained on the highly unbalanced dataset composed of the few exemplars from the previous task and fresh new samples from the new ones. As a result, the network is biased towards the data of new tasks, and this can lead to distorted predictions of the model, which is called \textit{task-recency bias}. To solve this problem, \emph{Bias Correction} (BiC) \cite{wu2019large} introduces a two-stage training where they perform the main training in the first stage and subsequently mitigate the bias through a linear transformation. Likewise, Weight Aligning (WA) \cite{zhao2020maintaining} proposes two-stage training. The first stage is equal to that of BiC and they normalize the weight vectors of the new classes and the old classes to reduce the bias in the second stage. Another bias correction method called \emph{Learning a Unified Classifier Incrementally via Rebalancing} (LUCIR) \cite{hou2019learning} alleviates task-recency bias by including three components into their training: cosine normalization, less-forget constraint, and inter-class. Built upon LUCIR, \emph{Pooled Outputs Distillation Network} (PODNet) \cite{douillard2020podnet} applies pooled out distillation loss and local similarity classifier.

\textbf{Dynamic Structure Method: } Dynamic structure approaches use masking for each task or expansion of the model to prevent forgetting and increase the model capacity to learn a new task. For instance, \emph{Conditional Channel Gated Networks} (CCGN)~\cite{abati2020conditional} dynamically adds an extra convolutional layer whenever the model learns a new task and it is only optimized for the new data. \emph{Adaptive Aggregation Networks} (AANets) \cite{liu2021adaptive} expands a Residual Network (ResNet) \cite{he2016deep} to have the two types of residual block at each residual level to balance stability and plasticity: a stable block that is trained on a first task and frozen afterward and a plastic block that is freely trained on a current task. Another dynamic structure method called \emph{Dynamically Expandable Representation Learning} (DER) \cite{yan2021dynamically} suggests to expand a feature extractor. The new feature extractor is trained solely on the current dataset with channel level masking and the whole model is fine-tuned on balanced dataset.

\section{Method}
\label{sec:method}

In this Section, we propose \textit{Auxiliary Network Continual Learning} (ANCL), a framework which combines original \textit{Continual Learning} (CL) approaches with an auxiliary network (\cref{subsec:formulation_of_ANCL}). In addition, we explain the detailed training steps of ANCL (\cref{subsec:algorithm_of_ANCL}).

\begin{figure*}
\centerline{\includegraphics[scale=.45]{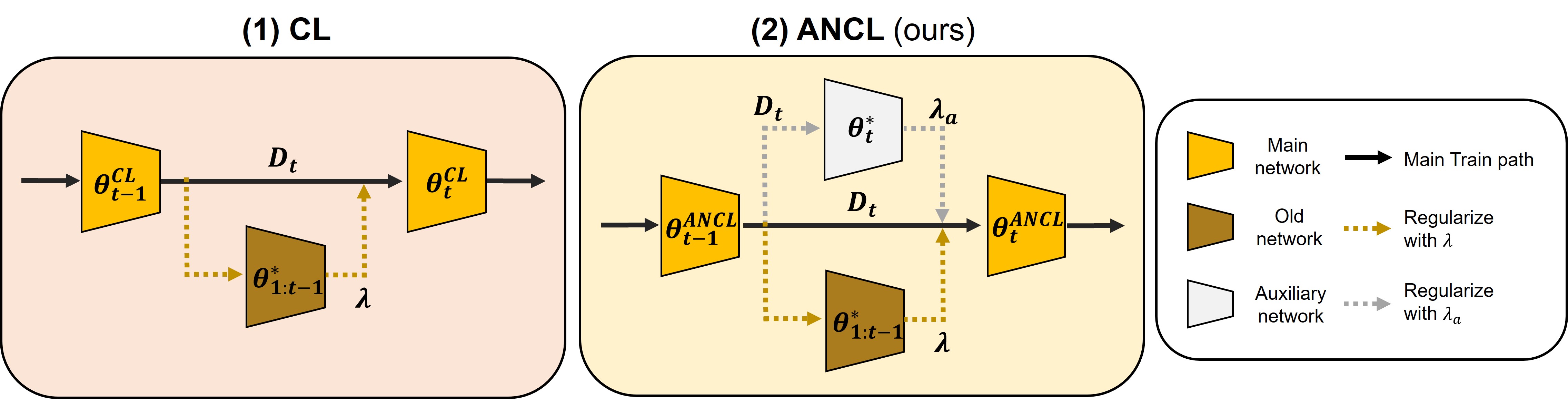}}
\caption{Conceptual comparison of \emph{Continual Learning} (CL) and \emph{Auxiliary Network Continual Learning} (ANCL) (ours) on task $t$. (1) CL: the previous weights $\theta^{CL}_{t-1}$ are frozen in the old network as $\theta_{1:t-1}^*$ and the old network regularizes the main training through $\lambda$. (2) ANCL: the auxiliary network initialized by $\theta_{t-1}^{ANCL}$ is trained on the dataset $D_t$ and then frozen as $\theta_t^*$. It regularizes the main training via $\lambda_a$ in addition to the regularization of the old network.}
\label{fig:CL_ANCL_concept}
\end{figure*}

\subsection{The Formulation of Auxiliary Network Continual Learning}\label{subsec:formulation_of_ANCL}

ANCL applies the auxiliary network trained on the current task to the continually learned previous network to achieve a balance between stability and plasticity. \cref{fig:CL_ANCL_concept} illustrates the conceptual difference between CL and ANCL, where CL can be any continual learning method that includes a regularizer that depends on the old network. Before training on the dataset $D_t$ of task $t$, CL freezes and copies the previous continual model $\theta^{CL}_{t-1}$ that has been trained until task $t-1$ as the old network $\theta_{1:t-1}^*$. Then, the old network regularizes the main training through the regularization strength $\lambda$. We can formally define the loss of CL on task $t$ as follows:
\begin{equation}
    \mathcal{L}_\text{CL} = \mathcal{L}_\text{t}(\theta) + \Omega(\theta ; \theta_{1:t-1}^*, \lambda), \label{eq:CL_loss}
\end{equation}
where the first term denotes a task-specific loss with respect to main network weights $\theta\in\mathbb{R}^P$ and the second term represents the regularizer that binds the dynamic of the network parameters $\theta$ to the old network parameters $\theta_{1:t-1}^*\in\mathbb{R}^P$. $\lambda\in\mathbb{R}$ is the regularization strength which is usually selected by a grid search procedure. These two loss terms can be calculated on the current dataset $D_t$ or on the combined dataset $D_t^+$ (current dataset $D_t$ + previous exemplars $P_{1:t-1}$) depending on the method to which it is applied. In classification problems, the task-specific loss becomes cross-entropy loss.  
The original CL approaches mainly focus on retaining the old knowledge obtained from the previous tasks by preventing large updates that would depart significantly from the old weights $\theta_{1:t-1}^*$. However, this might harmfully restrict the model's ability to learn the new knowledge, which will hinder the right balance between stability and plasticity. 

On the contrary, ANCL keeps two types of network to maintain this balance: (1) the auxiliary network $\theta_t^*$, which is solely optimized on the current task $t$ allowing for forgetting (\textit{plasticity}) and (2) the old network $\theta_{1:t-1}^*$ that has been sequentially trained until task $t-1$ (\textit{stability}). 
Then, both models are used to construct the regularizers in the following objective:
\begin{equation}
    \mathcal{L}_\text{ANCL} = \mathcal{L}_\text{t}(\theta) + \Omega(\theta ; \theta_{1:t-1}^*, \lambda) + \Omega(\theta ; \theta_t^*, \lambda_a)\label{eq:ANCL_loss} ,
\end{equation}
where the first two terms are the same as in \cref{eq:CL_loss} and the last term promotes the learning of the new task $t$ based on the parameters of the auxiliary network $\theta_t^*\in\mathbb{R}^P$ and the regularization strength $\lambda_a\in\mathbb{R}$. Note that the new regularizer $\Omega(\theta ; \theta_t^*, \lambda_a)$ is obtained in the same way as the original method, and thus we expect our model to naturally merge the old feature representation (or weight itself) with the new one. This is mathematically explained in \cref{Appendix:E} where we analyze and compare the gradient of CL and ANCL. 
Moreover, we initialize the auxiliary network with the old network parameters so that the auxiliary model is weakly biased toward the old model, thus facilitating the integration of the two models in the corresponding regularizers of Eq. \ref{eq:ANCL_loss}.

\begin{table}
\centering
\begin{tabular}{cc}
\hline
Methods                              & $\Omega(\theta ; \theta^*, \lambda)$         \\ \hline
EWC \cite{kirkpatrick2017overcoming} & $\frac{\lambda}{2}\sum_{i} F_{i}(\theta_{i}-\theta_{i}^*)^2$     \\
MAS \cite{aljundi2018memory}         & $\frac{\lambda}{2}\sum_{i} M_{i}(\theta_{i}-\theta_{i}^*)^2$     \\
LwF \cite{li2017learning}            & $\lambda \sum_{c=1}^{C_{1:t}} -y^c(x;\theta^*) \log{y^c(x;\theta)}$ \\
LFL \cite{jung2016less}              & $\lambda \norm{f(x; \theta) - f(x;\theta^*)}_2^2$  \\ \hline
iCaRL \cite{rebuffi2017icarl}        & $\lambda \sum_{c=1}^{C_{1:t}} -y^c(x;\theta^*) \log{y^c(x;\theta)}$ \\
BiC \cite{wu2019large}               & $\lambda \sum_{c=1}^{C_{1:t}} -y^c(x;\theta^*) \log{y^c(x;\theta)}$ \\
LUCIR \cite{hou2019learning}         &   $\lambda (1-\langle\bar{f}(x;\theta), \bar{f}(x;\theta^*)\rangle)$   \\
PODNet \cite{douillard2020podnet}    &  \begin{tabular}[r]{@{}r@{}} $\lambda [ \sum_{l=1}^{L-1} \mathcal{L}_\text{POD-spatial}(f_l(x;\theta), f_l(x;\theta^*))$ \\  $+ \mathcal{L}_\text{POD-flat}(f_L(x;\theta), f_L(x;\theta^*))]$ \end{tabular}   \\ \hline
\end{tabular}
\caption{The definition of $\Omega(\theta ; \theta^*, \lambda)$ depends on different methods. The first four methods (EWC, MAS, LwF, and LFL) are calculated on the current dataset $D_t$ while the last four methods (iCaRL, BiC, LUCIR, and PODNet) are measured on the combined dataset $D_t^+$ with the memory buffer. Detailed explanation and loss function of each method can be found in \cref{Appendix:B}.} 
\label{table:how_to_calculate_Omega}
\end{table}

In \cref{table:how_to_calculate_Omega}, we show how $\Omega(\theta ; \theta^*, \lambda)$ materializes in selected CL methods, given the current network parameters $\theta$, the reference network parameters $\theta^*$ (the old or auxiliary network), and the regularization strength $\lambda$. For example, the original CL loss of EWC can be expressed as follows by applying \cref{table:how_to_calculate_Omega} to \cref{eq:CL_loss}: 
\begin{equation}
    \mathcal{L}_\text{EWC} = \mathcal{L}_\text{t}(\theta) + \frac{\lambda}{2}\sum_{i} F_{1:t-1,i}(\theta_{i}-\theta_{1:t-1,i}^*)^2 \label{eq:ewc_loss}
\end{equation}
where $F_{t}$ is the approximation of the Fisher Information Matrix of the old network parameters $\theta_{1:t-1}^*$ and the regularization term calculates the difference between the network parameter $\theta_i$~($i = 1,\dots, P$) and the corresponding old network parameter $\theta_{1:t-1,i}^*$.
Next, if we apply ANCL to EWC to build the loss of the so-called \emph{Auxiliary Network EWC} (A-EWC), we get:
\begin{equation}
    \mathcal{L}_\text{A-EWC} = \mathcal{L}_\text{EWC}+ \frac{\lambda_a}{2}\sum_{i} F_{t,i}(\theta_{i}-\theta_{t,i}^*)^2\label{eq:a-ewc_loss}
\end{equation}
which adds the new regularizer built upon the auxiliary network parameter $\theta_{t,i}^*$. The application of ANCL to other methods in \cref{table:how_to_calculate_Omega} can be found in \cref{Appendix:C}.

In ANCL, the auxiliary network accounts for plasticity while the old network stands for stability. Furthermore, both networks are equally reflected through the regularization term $\Omega$, thus preventing bias toward either network. Adjusting both regularizers via $\lambda$ and $\lambda_a$, ANCL is more likely to achieve a better stability-plasticity balance than CL, under proper hyperparameter tuning. How ANCL solutions appropriately weigh the old network and the auxiliary network is further investigated in \cref{sec:trade-off_analysis}. Furthermore, we mathematically analyze and compare the gradient of CL and ANCL losses in terms of the stability-plasticity trade-off in \cref{Appendix:E}.

\paragraph{Comparison with AFEC} The auxiliary network of ANCL works similarly to the expanded parameter of AFEC with respect to adding an additional loss term, but ANCL uses a \textit{method-dependent} regularizer compared to the \textit{fixed and independent} regularizer of AFEC based on Fisher Information Matrix. In other words, while AFEC plugs in the same loss term calculated on the expanded parameter to every method, ANCL generates the loss term from the auxiliary network in the same way as the original CL where ANCL is applied. ANCL adopts two regularizers of the same type to equally represent stability and plasticity which is explicitly controlled by the scaling hyperparameters ($\lambda$ and $\lambda_a$ in \cref{eq:ANCL_loss}). If the two regularizers are of different types like in AFEC, each regularizer will change in different magnitude at every epoch. Consequently, it is less likely that the model will arrive at the best equilibrium. In \cref{Appendix:D}, we empirically show that ANCL outperforms AFEC.

\subsection{Algorithm} \label{subsec:algorithm_of_ANCL}
Detailed training steps of our ANCL framework is summarized in Alg.\ \ref{alg:DCL}. This is applicable to all ANCL methods if an appropriate ANCL loss is substituted in the algorithm. Given the training over total $N$ tasks, Lines 3-4 shows the training of the main network weight $\theta$ with task-specific loss $\mathcal{L}_\text{t}$ on the dataset of task $1$. Then, the optimal weight $\theta^*$ for task $1$ is saved as the old weight $\theta_{1:1}^*$ in Line 5. On task $t (>1)$, the auxiliary weight $\theta_t$ is initialized by the previous old weight $\theta_{1:t-1}^*$ and trained with task-specific loss $\mathcal{L}_\text{t}$ (Lines 7-9). In Line 10, the auxiliary weight $\theta_t^*$ is frozen and saved. Subsequently, the main network is trained with ANCL loss explained in \cref{eq:ANCL_loss} (Lines 11-12). The optimal main network on task $t$ is frozen and saved as an old weight $\theta_{1:t-1}^*$ for the next loop (Line 13). If Lines 7-10 are skipped and "ANCL Loss (\cref{eq:ANCL_loss})" in Line 12 is replaced with "CL Loss (\cref{eq:CL_loss})", Alg.\ \ref{alg:DCL} becomes the original CL algorithm. 

\begin{algorithm}[!ht]  
\DontPrintSemicolon 
  \KwInput{Main network weight $\theta$, Auxiliary network weight $\theta_t^*$, Old network weight $\theta_{1:t-1}^*$, Hyperparameters $\lambda$, $\lambda_a$}
  \KwOutput{Optimal main network weight $\theta^*$}
    \For{task t = 1, 2, .., N}    
        { \If{$t = 1$}
            {
                \tcp{Train main network}
                \For{epoch e = 1, 2, .., E} 
                {
                Train $\theta$ with task-specific loss $\mathcal{L}_\text{t}$ to obtain $\theta^*$ on task 1
            	}   
                \tcp{Save main network weight as old network weight}
                Freeze and save $\theta^*$ as $\theta_{1:1}^*$   
            } 
        
          \Else
            {
            \tcp{Initialize auxiliary network}
            $\theta_t$ = copy($\theta_{1:t-1}^*$) \\
            \tcp{Train auxiliary network}
            \For{epoch e = 1, 2, .., E} 
            {
            Train $\theta_t$ with task-specific loss $\mathcal{L}_\text{t}$ to obtain $\theta_t^*$ on task $t$ } 
            \tcp{Save auxiliary network weight}
            Freeze and save $\theta_t^*$ \\
            \tcp{Train main network}
            \For{epoch e = 1, 2, .., E} 
            {
            Train $\theta$ with ANCL Loss (\cref{eq:ANCL_loss}) to obtain $\theta^*$ on task $t$
        	} 
            \tcp{Save main network weight as old network weight}
            Freeze and save $\theta^*$ as $\theta_{1:t-1}^*$ 
            }
        }  
        \caption{ANCL Algorithm} \label{alg:DCL}
\end{algorithm}

\section{Experiment}
\label{sec:experiment}

\begin{table*}
\centering
\begin{tabular}{ccccc}
\hline
                                          & \multicolumn{2}{c}{CIFAR-100}               & \multicolumn{2}{c}{Tiny ImageNet}          \\
Methods                                   & (1)               & (2)                  & (3)                    & (4)                   \\ \hline
Fine-tuning                               & $38.90_{\pm1.59}$    & $27.81_{\pm0.80}$    & $28.51_{\pm0.75}$     & $20.35_{\pm1.70}$     \\
Joint                                     & $89.64_{\pm0.37}$    & $93.42_{\pm0.27}$    & $67.98_{\pm1.15}$     & $70.02_{\pm2.63}$     \\
LwM \cite{dhar2019learning}                & $78.46_{\pm1.11}$    & $78.27_{\pm0.38}$    & $59.04_{\pm0.63}$     & $59.78_{\pm1.08}$     \\
DMC \cite{zhang2020class}                  & $51.90_{\pm0.91}$    & $53.72_{\pm1.11}$    & $45.65_{\pm0.15}$     & $44.50_{\pm0.73}$     \\
\hline
EWC \cite{kirkpatrick2017overcoming}       & $58.13_{\pm0.87}$    & $60.03_{\pm1.23}$    & $50.10_{\pm0.78}$     & $52.53_{\pm0.91}$     \\
\rowcolor{Gray}
w/ ANCL (ours)                             & $60.86_{\pm1.46}$    & $62.47_{\pm0.65}$    & $52.49_{\pm0.71}$     & $53.86_{\pm0.88}$     \\ \hline
MAS \cite{aljundi2018memory}               & $60.56_{\pm0.82}$    & $59.35_{\pm1.09}$    & $49.50_{\pm1.18}$     & $51.79_{\pm0.51}$     \\
\rowcolor{Gray}
w/ ANCL (ours)                             & $64.43_{\pm1.17}$    & $60.70_{\pm1.11}$    & $50.11_{\pm1.09}$     & $53.58_{\pm0.73}$     \\ \hline
LwF \cite{li2017learning}                  & $78.87_{\pm0.69}$    & $76.96_{\pm0.83}$    & $59.04_{\pm0.62}$     & $62.09_{\pm0.59}$     \\
\rowcolor{Gray}
w/ ANCL (ours)                             & $79.42_{\pm0.57}$ & $79.99_{\pm0.59}$ & $60.96_{\pm0.76}$     & $63.79_{\pm0.41}$     \\ \hline
LFL \cite{jung2016less}                    & $74.50_{\pm0.57}$    & $74.27_{\pm0.72}$    & $60.20_{\pm0.66}$     & $58.47_{\pm0.95}$     \\
\rowcolor{Gray}
w/ ANCL (ours)                             & $75.23_{\pm0.67}$    & $74.68_{\pm1.04}$    & $61.32_{\pm0.68}$     & $58.98_{\pm0.74}$     \\ \hline
\end{tabular}
\caption{The averaged accuracy (\%) on the benchmarks (1)-(4). Reported metrics are averaged over 3 runs (averaged accuracy $\pm$ standard error). ANCL methods are colored gray.} 
\label{table:mean_acc}
\end{table*}

\begin{table*}
\centering
\begin{tabular}{ccccc}
\hline
                              & \multicolumn{2}{c}{CIFAR-100}          & \multicolumn{2}{c}{Tiny ImageNet} \\
Methods                       & (5)                  & (6)                     & (7)                   & (8)               \\ \hline
Fine-tuning                   & $45.78_{\pm0.90}$    & $43.57_{\pm1.33}$    & $27.44_{\pm0.85}$     & $24.18_{\pm0.98}$     \\
Joint                         & $67.84_{\pm1.35}$    & $66.40_{\pm0.86}$    & $46.85_{\pm0.74}$     & $46.02_{\pm0.55}$     \\
EEIL \cite{castro2018end}     & $49.81_{\pm1.12}$     & $48.65_{\pm0.94}$     & $28.68_{\pm0.93}$     & $28.00_{\pm0.73}$ \\ \hline
iCaRL \cite{rebuffi2017icarl} & $58.05_{\pm0.94}$         & $57.11_{\pm0.77}$        & $39.04_{\pm0.61}$             & $37.90_{\pm0.98}$          \\
\rowcolor{Gray}
w/ ANCL (ours)             & $61.22_{\pm0.88}$         & $59.13_{\pm0.68}$        & $41.46_{\pm0.85}$             & $39.91_{\pm1.02}$        \\
BiC \cite{wu2019large}      & $56.74_{\pm1.33}$         & $55.73_{\pm1.21}$        & $40.56_{\pm0.44}$             & $39.21_{\pm0.69}$      \\
\rowcolor{Gray}
w/ ANCL (ours)                    & $58.32_{\pm1.27}$         & $58.23_{\pm1.44}$        & $42.61_{\pm0.65}$             & $40.56_{\pm0.51}$        \\ \hline
LUCIR \cite{hou2019learning}     & $56.06_{\pm0.45}$         & $57.91_{\pm0.57}$        & $35.17_{\pm0.58}$             & $30.02_{\pm0.13}$         \\
\rowcolor{Gray}
w/ ANCL (ours)                    & $60.20_{\pm0.78}$         & $60.04_{\pm0.80}$        & $37.89_{\pm0.74}$             & $31.65_{\pm0.25}$         \\ \hline
PODNet \cite{douillard2020podnet} & $61.80_{\pm0.77}$         & $59.22_{\pm0.93}$        & $40.28_{\pm0.36}$             & $38.50_{\pm0.49}$            \\
\rowcolor{Gray}
w/ ANCL (ours)                    & $63.15_{\pm0.62}$         & $60.44_{\pm0.67}$        & $41.11_{\pm0.23}$             & $40.11_{\pm0.64}$           \\ \hline
\end{tabular}
\caption{The averaged incremental accuracy (\%) on the benchmarks (5)-(8). Reported metrics are averaged over 3 runs (averaged accuracy $\pm$ standard error). ANCL methods are colored gray.} 
\label{table:mean_acc_replay}
\end{table*}

\textbf{Benchmark:} CIFAR-100 \cite{krizhevsky2009learning} and Tiny ImageNet \cite{le2015tiny} are chosen to evaluate ANCL. CIFAR-100 contains 60,000 colored images from 100 classes with the size of $32 \times 32$. For task incremental scenario, CIFAR-100 is divided into 10 tasks of 10 classes each and 20 tasks of 5 classes each to construct two benchmarks: \textbf{(1) CIFAR-100/10} and \textbf{(2) CIFAR-100/20}. In addition, we build two more benchmarks for class incremental scenario: \textbf{(5) CIFAR-100/6} and \textbf{(6) CIFAR-100/11}. In these settings, 50 classes are learned at an initial phase and the rest classes are learned sequentially with 10 classes or 5 classes per phase after the initial one. Tiny ImageNet consists of 110,000 colored images (size $64 \times 64$) from 200 classes which are resized as $32 \times 32$ for both training and inference. We equally divide Tiny ImageNet into 10 and 20 tasks to build two benchmarks for task incremental scenario: \textbf{(3) TinyImagenet-200/10} and \textbf{(4) TinyImagenet-200/20}. For class incremental scenario, the model is trained on 100 classes at an initial phase and then trained continuously on 10 classes or 5 classes per phase after the initial one: \textbf{(7) TinyImagenet-200/11} and \textbf{(8) TinyImagenet-200/21}. 

\textbf{Architecture:} We select Resnet32 \cite{he2016deep} for all benchmarks which is commonly chosen in the literature of continual learning \cite{rebuffi2017icarl,wu2019large, hou2019learning,zhao2020maintaining,douillard2020podnet}. For task incremental scenario, multi-head layer is deployed instead of the last layer in Resnet32 to generate an output with a task identity. In class incremental scenario, single-head evaluation is adopted due to the absence of task identity during inference. 

\textbf{Implementation:} The model is trained from scratch and every experiment is carried out 3 times with different seeds to generate averaged metrics. SGD optimizer with momentum 0.9 and batch size 128 is applied to all experiments. In task incremental learning, we evaluate our methods on a strict setting of continual learning where the previous data is not visited again. In class incremental learning, we relax the regularization of accessing previous data. 20 exemplars per class of the old training data are selected by herding sampling strategy and stored in the memory buffer (more details in \cref{Appendix:F.1}). 

\textbf{Gridsearch on Parameters:} We conduct a comprehensive hyperparameter search for all methods and report the best scores for a fair comparison. We follow the way AFEC \cite{wang2021afec} performs the grid search on $\lambda$ and $\lambda_a$. First, an extensive grid search is made on $\lambda$ using the original CL loss and $\lambda$ is fixed afterward. Then, we use ANCL loss to conduct the grid search of $\lambda_a$. Grid search result of $\lambda$ and $\lambda_a$ for all benchmarks can be found in \cref{Appendix:F.4}

\textbf{Evaluation Metrics:} In task incremental scenario, averaged accuracy ($AAC$) for $T$ task is calculated after the training of all tasks. In class incremental scenario, averaged incremental accuracy ($AIAC$) is used instead:
\begin{equation}
    AAC = \frac{1}{T}\sum_{i=1}^T A_{T,i}, \quad AIAC = \frac{1}{N+1}\sum_{i=0}^N A_{i}.
\end{equation}
In AAC, $A_{j,k}$ is the test accuracy of task $k$ after the continual learning of task $j$. In AIAC, $A_i$ denotes the test accuracy of the classes seen so far at the $i$th phase for the benchmark consisting of $N+1$ phases including the initial one.

\begin{figure*}[htbp]
\centerline{\includegraphics[scale=.45]{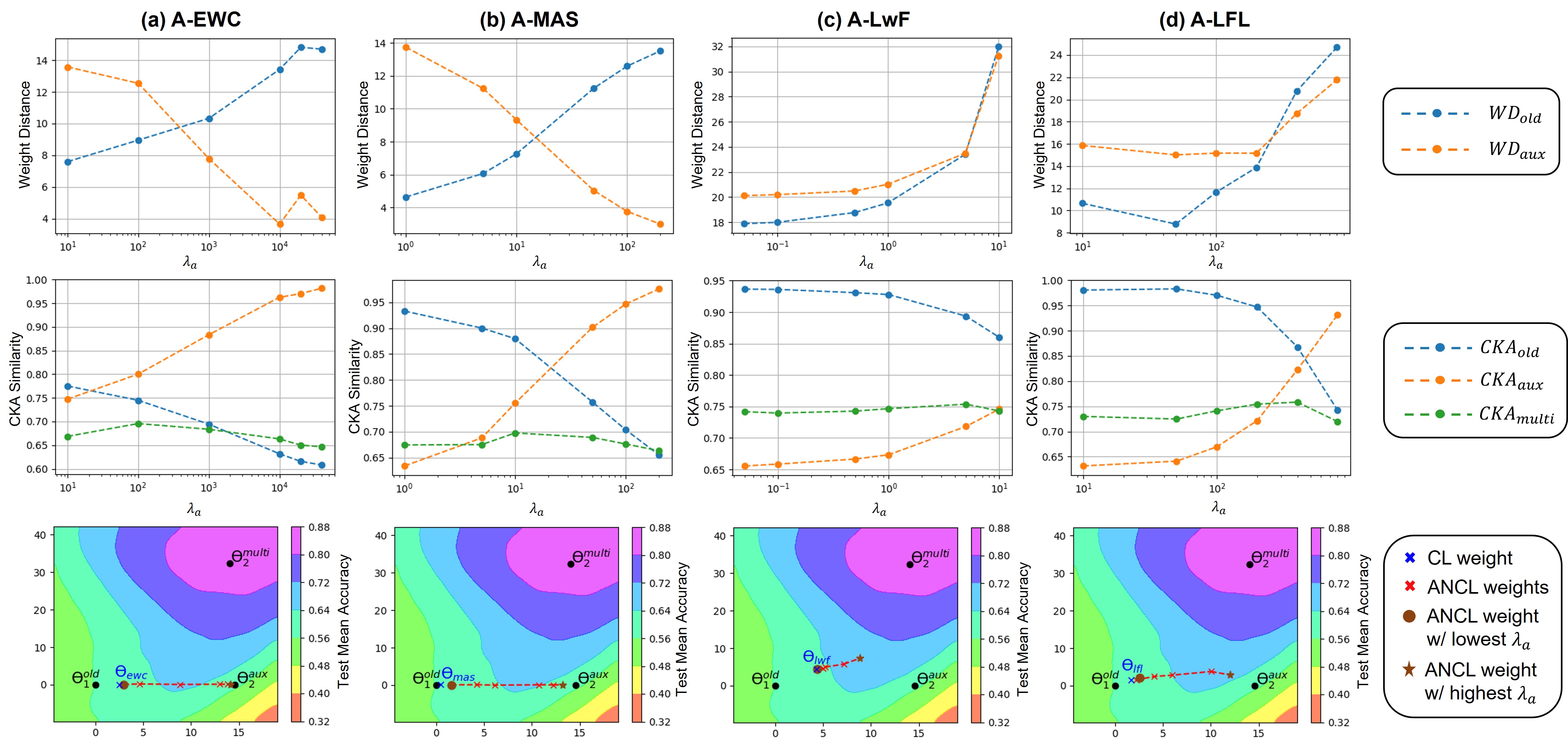}}
\caption{Anaylsis figures on (1) CIFAR-100/10: weight distance (top row), centered kernel alignment (middle row), and mean accuracy landscape (bottom row). The set of $\lambda_a$ for each ANCL is as follows ($\lambda$ is fixed): (a) A-EWC ($\lambda=10000$) - $\lambda_a \in$ [10, 100, 1000, 10000, 20000, 40000], (b) A-MAS ($\lambda=50$) - $\lambda_a \in$ [1, 5, 10, 50, 100, 200], (c) A-LwF ($\lambda=10$) - $\lambda_a \in$ [0.05, 0.1, 0.5, 1, 5, 10] and (d) A-LFL ($\lambda=400$) - $\lambda_a \in$ [10, 50, 100, 200, 400, 800].}
\label{fig:trade-off_figures}
\end{figure*}

\textbf{Baseline:} Fine-tuning is the naive approach that a model is fine-tuned on each task (or each phase), which is regarded as a lowerbound and joint uses the whole dataset to train the model, which becomes an upperbound. In task incremental setting, we evaluate EWC \cite{kirkpatrick2017overcoming}, MAS \cite{aljundi2018memory}, LwF \cite{li2017learning}, LFL \cite{jung2016less}, LwM \cite{dhar2019learning}, and DMC \cite{zhang2020class}. For a fair comparison, DMC is modified to only use the original dataset like other methods instead of an unlabeled auxiliary dataset. Then, we apply ANCL to the original CL approaches. 
In class incremental setting, we test EEIL \cite{castro2018end}, iCaRL \cite{rebuffi2017icarl}, BiC \cite{wu2019large}, LUCIR \cite{hou2019learning}, and PODNet \cite{douillard2020podnet} with their applications to ANCL.

\textbf{Evaluation on Task Incremental Scenario:} \cref{table:mean_acc} shows that applying ANCL consistently gives an extra boost in accuracy by 1-3 \% compared to naive CL and A-LwF achieves the best accuracy in all benchmarks. ANCL can be more compatible with specific methods than others. For example in benchmark (1), applying ANCL outperforms MAS baseline by 3.87 \% while it improves LFL baseline only by 0.73 \%. This is because ANCL is more effective when the two regularizers in \cref{eq:ANCL_loss} are well suited to each other and CL has less plasticity at the beginning. The detail accuracy for all tasks can be found in \cref{Appendix:F.2}.

\textbf{Evaluation on Class Incremental Scenario:} In \cref{table:mean_acc_replay}, we can clearly see that ANCL surpasses CL baselines in all methods by 1-3 \% including state-of-the-art (SOTA) methods such as BiC \cite{wu2019large}, LUCIR \cite{hou2019learning}, and PODNet \cite{douillard2020podnet}. Similarly to \cref{table:mean_acc}, ANCL is more compatible with LUCIR and iCaRL compared to others thereby A-iCaRL being able to compete with or even outperform the stronger baseline of PODNet. We also plot how each method's accuracy at each phase changes and report the final accuracy in \cref{Appendix:F.3}.

\section{Stability-Plasticity Trade-off Analysis}\label{sec:trade-off_analysis}

In this chapter, we perform three analyses on (1) CIFAR-100/10 to study how the stability-plasticity dilemma is solved through ANCL: \emph{Weight Distance}, \emph{Centered Kernel Alignment}, and \emph{Mean Accuracy Landscape}. For simplification, $\lambda$ is first selected by grid search using CL loss on current task $t=2$ and then fixed. Then, ANCL solutions with different $\lambda_a$ are compared in various analyses. A training regime similar to the one in \cite{mirzadeh2020linear} is adopted for a fair comparison, which is explained in detail in \cref{Appendix:G.1}.

\subsection{Weight Distance} \label{subsec:weigth_distance_analysis}
If the parameters change less, it is reasonable to expect that less forgetting will occur. According to \cite{mirzadeh2020understanding}, forgetting $\mathcal{F}_1$ on task $1$ is bounded using Taylor expansion of the loss as follows:
\begin{align}
    \mathcal{F}_1 &= \mathcal{L}_1(\hat{\theta}_2) - \mathcal{L}_1(\hat{\theta}_1) \\&\approx \frac{1}{2} (\hat{\theta}_2 - \hat{\theta}_1)^T \nabla^2 \mathcal{L}_1(\hat{\theta}_1) (\hat{\theta}_2 - \hat{\theta}_1) \\ 
    &\le \frac{1}{2} \lambda_1^{max} \norm{\hat{\theta}_2 - \hat{\theta}_1}_2^2 
\end{align}
where $\mathcal{L}_1$ is the empirical loss on task $1$ and $\nabla^2 \mathcal{L}_1(\hat{\theta}_1)$ is the Hessian for $\mathcal{L}_1$ at $\hat{\theta}_1$. $\lambda_1^{max}$ is the maximum eigenvalue of $\nabla^2 \mathcal{L}_1(\hat{\theta}_1)$. Above inequality implies that the bound of forgetting $\mathcal{F}_1$ is determined by the norm of the difference between two weights near the minima of task 1 loss.

On task $t$, we measure the weight distance (WD) from the weights of the ANCL models $\theta_t^{ANCL}$ to the weights of the old model $\theta_{t-1}^{old}$ and the auxiliary model $\theta_{t}^{aux}$ respectively:
\begin{align}
    WD_{old} &= \norm{\theta_t^{ANCL} - \theta_{t-1}^{old}}_2, \\
    WD_{aux} &= \norm{\theta_t^{ANCL} - \theta_{t}^{aux}}_2.
    \label{eq:weight_distance_def}
\end{align}
WD analysis is shown in the top row of \cref{fig:trade-off_figures}. We calculate WD with different $\lambda_a$ which directly adjusts the stability-plasticity trade-off while $\lambda$ is fixed. The model parameters remain close to the old parameters when $\lambda_a$ is small, which can be seen on the left side of all WD figures. For A-EWC and A-MAS, $WD_{aux}$ decreases and $WD_{old}$ increases as $\lambda_a$ becomes larger. This result implies a direct interpolation between the old and auxiliary networks, which is consistent with the analysis of the ANCL gradient in \cref{Appendix:E}. For A-LwF and A-LFL, $WD_{aux}$ becomes relatively smaller than $WD_{old}$ with increasing $\lambda_a$ but $WD_{old}$ and $WD_{aux}$ are both growing. Unlike EWC and MAS which directly regularize the weights itself, LwF and LFL have more flexibility to remember the previous knowledge by utilizing loss terms based on activations or logits. Therefore, for the distillation approaches, the model weights tends to move relatively closer to the auxiliary weights with increasing $\lambda_a$ but not directly toward it like EWC or MAS. The difference between the regularization and distillation CL methods and the effect of $\lambda_a$ on the stability-plasticity trade-off is studied further in the following analyses.

\subsection{Centered Kernel Alignment}\label{subsec:CKA_analysis}
Centered Kernel Alignment (CKA) \cite{kornblith2019similarity} measures the similarity of two-layer representations on the same set of data. Given $N$ data and $p$ neurons, the layer activation matrices $R_1 \in \mathbb{R}^{N \times p}$ and $R_2 \in \mathbb{R}^{N \times p}$ are generated by two layers from two independent networks. Then, CKA is defined as:
\begin{align}
    CKA(R_1, R_2) = \frac{HSIC(R_1, R_2)}{\sqrt{HSIC(R_1, R_1)}\sqrt{HSIC(R_2, R_2)}}
\end{align}
where $HSIC$ stands for Hilbert-Schmidt Independence Criterion \cite{gretton2005measuring}. We use linear $HSIC$ to implement CKA. 
It is well known that lower layers have relatively higher CKA scores than deeper layers and deeper layers generally contribute to forgetting \cite{ramasesh2020anatomy}. In this analysis, we measure three CKA similarity:
\begin{align}
    CKA_{old} &= \frac{1}{L}\sum_{l=1}^L CKA(R_{t,l}^{ANCL}, R_{t-1,l}^{old}),\label{eq:CKA_def1}\\
    CKA_{aux} &= \frac{1}{L}\sum_{l=1}^L CKA(R_{t,l}^{ANCL}, R_{t,l}^{aux}), \label{eq:CKA_def2}\\
    CKA_{multi} &= \frac{1}{L}\sum_{l=1}^L CKA(R_{t,l}^{ANCL}, R_{t,l}^{multi}). \label{eq:CKA_def3}
\end{align}
where CKA is calculated and averaged over the set of layers $\{1, \dots, L\}$ in Resnet32. Resnet32 consists of 1 initial convolution layer and 3 residual blocks. In order to measure the output similarity of two networks, we select 10 convolution layers in the last residual block of Resnet32 as our set. $R_{t}^{ANCL}$, $R_{t-1}^{old}$ and $R_{t}^{aux}$ are the activation matrices of the ANCL network, the old network, and the auxiliary network, respectively. $R_{t}^{multi}$ is the activation output of the multitask model trained on the entire dataset $D_{1:t}$ until the task $t$. If $CKA_{multi}$ is high, the model generates layer activations similar to those of the multitask model. Then, the model is highly likely to perform well on all tasks like the multitask model, which is the main goal of continual learning.

The middle row of \cref{fig:trade-off_figures} shows three CKA similarities with different $\lambda_a$. In all methods, increasing $\lambda_a$ results in higher $CKA_{aux}$ and lower $CKA_{old}$, which can be interpreted to mean that the representations of the ANCL network become more similar to that of the auxiliary network and less similar to that of the old network. We can clearly see that the stability-plasticity trade-off is controlled by $\lambda_a$ through the interaction between the old and auxiliary networks. On the other hand, if $CKA_{multi}$ reaches the highest score at specific $\lambda_a$, that model is highly likely to have the best trade-off. For example, (b) A-MAS and (d) A-LFL achieve the highest $CKA_{multi}$ at $\lambda_a = 10$ and $\lambda_a = 400$ respectively. In general, $CKA_{multi}$ of the distillation methods is higher than that of the regularization methods, which corresponds to the results in \cref{table:mean_acc} where the distillation methods achieved a higher averaged accuracy compared to the regularization methods.

\subsection{Mean Accuracy Landscape}\label{subsec:mean_acc_landscape_analysis}

Lastly, we visualize mean accuracy landscape of task $1$ and $2$ in weight vector space following \cite{mirzadeh2020linear} (details in \cref{Appendix:G.3}). $\theta_1^{old}$, $\theta_2^{aux}$, and $\theta_2^{multi}$ are used to build two-dimensional subspace denoting the weights of the old network, the auxiliary network and the multitask network, respectively. Multitask network is trained on whole dataset $D_{1:2}$ until task $2$ and thus $\theta_2^{multi}$ is located in the highest contour indicating the highest mean accuracy. We project CL (blue) and ANCL (red) weight vectors on the subspace to see how ANCL parameters are shifted on the accuracy landscape with different $\lambda_a$. ANCL weights with the lowest $\lambda_a$ are denoted as a brown circle and $\lambda_a$ increases following the red dot line. Finally, the red dot line reaches a brown star which indicates ANCL weights with the highest $\lambda_a$.

In A-EWC and A-MAS, it is clearly observed that $\lambda_a$ adjusts the interpolation between the CL weights $\theta_{CL}$ and the auxiliary weights $\theta_2^{aux}$. The large $\lambda_a$ drifts the ANCL weights $\theta_{ANCL}$ directly toward $\theta_2^{aux}$ and the ANCL with sufficiently small $\lambda_a$ converges to CL methods. At the interpolation of the old weights $\theta_1^{old}$ and the auxiliary weights $\theta_2^{aux}$, the ANCL weight achieves higher mean accuracy located in the higher contour. Similarly in A-LwF and A-LFL, $\theta_{ANCL}$ with the lowest $\lambda_a$ starts near $\theta_{CL}$ and tends to move toward the region between $\theta_1^{old}$ and $\theta_2^{aux}$. As the distillation methods have more flexibility to retain the previous knowledge, the weights of A-LwF and A-LFL do not directly move toward $\theta_2^{aux}$ like those of A-EWC and A-MAS. Because of its flexibility, ANCL with distillation methods can deviate from the interpolation line and climb to the higher contour of mean accuracy. As a result, the best trade-off is made at somewhere between $\theta_1^{old}$ and $\theta_2^{aux}$. Again, the mean accuracy landscape figures show the projection of weight in the two-dimensional subspace built by three weights ($\theta_1^{old}$, $\theta_2^{aux}$, and $\theta_2^{multi}$). Therefore, it approximates the relative positions of CL and ANCL weights but does not reflect the exact positions of them in the weight space.

As a result, three analyses strongly support the notion that ANCL is able to achieve a better stability-plasticity trade-off where $CKA_{multi}$ and mean accuracy are the highest. The trade-off is mainly adjusted by the ratio between $\lambda$ and $\lambda_a$. ANCL with high $\lambda_a$ infuses more plasticity into the model, while ANCL with low $\lambda_a$ seeks more stability. These results coincides with the analysis of ANCL in \cref{Appendix:E} where the solutions of A-EWC and A-MAS indicate the explicit interpolation between the old and auxiliary weights and the gradients of A-LwF and A-LFL derive the activation (or logit) of the main network toward the interpolated activation (or logit) between the old and auxiliary networks. 

\section{Conclusion}
\label{sec:conclusion}
In our paper, we propose a novel framework called ANCL to pursue the proper balance between stability and plasticity inspired by the recent works \cite{wang2021afec, zhang2020class, liu2021adaptive, lin2022towards} adopting an auxiliary network. Our method outperforms the original baselines, including SOTA methods on CIFAR-100 \cite{krizhevsky2009learning} and Tiny ImageNet \cite{le2015tiny}. To investigate the underlying mechanism of ANCL, we extensively conduct analyses and confirm that the balance is resolved via the interpolation between the old and auxiliary weights. In summary, our work provides a deeper understanding of the interaction between the old network and the auxiliary network, which is the key to recent research on continual learning. 

Although ANCL can achieve better stability-plasticity trade-off compare to CL, it should be supported by enough hyperparameter search of $\lambda$ and $\lambda_a$. Therefore, extra computational burdens are required to search appropriate hyperparameters for each method, and results can be variant depending on the scope of grid search. In the future, we will investigate a better way to find these hyperparameters such as in data-driven fashion or inside the optimization process.

{\small
\bibliographystyle{ieee_fullname}
\bibliography{PaperForReview}
}

\onecolumn
\newpage

\appendix

\section{Motivation of Our Work: Using Auxiliary Network for Stability-Plasticity Balance}\label{Appendix:A}
In the last few years, several papers~\cite{wang2021afec, zhang2020class, liu2021adaptive, lin2022towards} have proposed to use an auxiliary network or an extra module which is solely trained on current dataset. They tried to combine this additional structure with a previous network or module which has been trained continually on old datasets. 

\emph{Elastic Weight Consolidation} (EWC) \cite{kirkpatrick2017overcoming} is one of the initial works among weight regularization methods that regularize weights related to the previous tasks. After training each task, EWC copies and freezes the current network as an old network. Then, it estimates the importance of each parameter in the old network so that the weights with the high importance remain unchanged when the model is optimized on future task. Based on the EWC, \cite{wang2021afec} proposes \emph{Active Forgetting with synaptic Expansion-Convergence} (AFEC) which further regularizes the weights relevant to the current task through a new set of parameters called expanded parameters. The expanded parameters are solely trained on the dataset of the new task initialized by the old network and are allowed to forget the previous tasks. As a result, AFEC can reduce potential negative transfer in continual learning by selectively merging the old parameters with the expanded parameters. The stability-plasticity balance in AFEC is adjusted via hyperparameters which scale the regularization terms for remembering the old tasks and learning the new tasks.

\emph{Learning without Forgetting} (LwF) \cite{li2017learning} prevents forgetting using knowledge distillation \cite{bucilua2006model, hinton2015distilling}. They apply a distillation loss so that the model can learn soft targets generated by the old model instead of typical one-hot targets. The old model is copied and freezed before the training of the current task like EWC. \emph{Deep Model Consolidation} (DMC) \cite{zhang2020class} is another distillation method built upon LwF. DMC proposes double distillation loss where the soft targets are generated using both the old model and a new model. The new model in DMC is basically the same concept as the expanded parameters in AFEC which is optimized on the current task. Then, the logits of new classes from the new model and the logits of old classes from the old model are concatenated to build the final soft targets. Then, the stability-plasticity balance is achieved by penalizing the model to generate the same output as the old model for old classes and the same output as the new model for new classes. 

Adaptive Aggregation Networks (AANet) \cite{liu2021adaptive} explicitly expands ResNet \cite{he2016deep} to have the two types of residual blocks at each residual level: the one for retaining old knowledge and the other for learning new knowledge. The outputs from the two residual blocks are linearly combined by aggregation weights and then proceeded to the next-level layer. They train AANets through bilevel optimization. In the first level, the parameters of the two residual blocks are trained. In the second level, the aggregation weights are adapted which decide the balance between stability and plasticity within the ResNet.

Recent work by \cite{mirzadeh2020linear} observes that multitask and continual solutions are connected by very simple curves with a low error in weight space, which is called \emph{Linear Mode Connectivity}. They empirically prove that this connectivity is a linear path if the multitask learning and the continual learning share same initialization weights. Based on this observation, \cite{lin2022towards} proposes a simple linear connector that linearly add the weights of two networks following this linear path to emulate the multitask solution: the one model remembering the old tasks and the other model learning the new tasks. The stability-plasticity balance is preserved by combining the two networks.

The above methods~\cite{wang2021afec, zhang2020class, liu2021adaptive, lin2022towards} all share the property that the auxiliary model or module is used to solve the stability-plasticity dilemma in continual learning. Consequently, these methods are able to learn the current task better than the original method while still retaining the knowledge of the previous tasks. However, the underlying mechanism of the interaction between the previous model and the auxiliary model is not widely studied. Therefore, in this work, we first formalize the framework of continual learning that adopts the auxiliary network called \emph{Auxiliary Network Continual Learning} (ANCL). Given this environment, we investigate the stability-plasticity trade-off from both a theoretical and empirical point of view and perform various analyses to better understand it.

\section{The Detail Explanation of Methods in \cref{table:how_to_calculate_Omega} of Main Paper}\label{Appendix:B}

\subsection{Weight Regularization Method}
The continual learning aims to learn sequentially $T$ tasks using a neural network $f_\theta$, where $\theta\in\mathbb{R}^P$ denotes the learnable weights. In the standard continual learning framework, when presented with task $t$, the user has an access to previous network weights $\theta_{1:t-1}^{*}\in\mathbb{R}^P$, which are the result of continual learning from task $1$ to $t-1$. It is well known that simply starting optimization from the weights $\theta_{1:t-1}^{*}$ to obtain the weights $\theta_{1:t}^{*}$ using the new data from the task $t$ results in catastrophic forgetting~\cite{mccloskey1989catastrophic} of the old tasks. A standard way to mitigate catastrophic forgetting is to include a regularization term which binds the dynamics of each network parameter $\theta_i$~($i\in 1,\dots, P$) to the corresponding old network parameter $\theta_{1:t-1,i}^*$ through a regularization term $R_{1:t-1,i}>0$. The new optimization problem on task $t$ then returns:
\begin{equation}
    \theta_{1:t}^* = \argmin_{\theta = (\theta_1,\dots,\theta_P)} \left[L_{\text{reg}} = \mathcal{L}_\text{t}(\theta) + \frac{\lambda}{2}\sum_{i} R_{1:t-1,i}(\theta_{i}-\theta_{1:t-1,i}^{*})^2\right]
    \label{eq:reg_loss}
\end{equation}
where in classification problems $\mathcal{L}_\text{t}(\theta)$ is cross-entropy loss on the data of the task $t$ and $\lambda$ is the regularization strength which is usually selected by a grid search procedure. $R_{1:t-1,i}$ is the accumulated regularizer of each parameter $\theta_{i}$ until task $t-1$ and various regularization-based methods choose different ways to estimate this parameter. 

For example, \emph{Elastic Weight Consolidation} (EWC) \cite{kirkpatrick2017overcoming} calculates $R_i$ through the approximation of Fisher Information Matrix (FIM). The diagonal elements of FIM quantifies how curved the likelihood of each parameter is and can be approximated by a Hessian matrix near the optimum. In other words, the larger the change in the gradient, the more relevant the corresponding parameter is to the previous tasks. FIM is calculated after training each task, which means that $R_i$ of EWC cannot fully reflect the learning trajectory of each network weight.

Compared to EWC, \emph{Memory Aware Synapses} (MAS) \cite{aljundi2018memory} proposes accumulating the changes of each parameter throughout the update history. $R_i$ is measured through the magnitude of the updates on each parameter according to the change in output. In other words, if the small change of specific parameter in a network causes the huge change of an output, that parameter should be memorized first.

\subsection{Knowledge Distillation Method}

Distillation-based approaches prevent forgetting through knowledge distillation \cite{bucilua2006model, hinton2015distilling} which was originally designed to train a more compact student network from a larger teacher network. In this way, the main network can emulate the activation or logit of the previous network while learning a new task.

\textit{Learning without Forgetting} (LwF) \cite{li2017learning} proposes the following loss on task $t$ to retain the old knowledge:
\begin{equation}
    \mathcal{L}_\text{LwF} = \mathcal{L}_\text{t}(\theta) + \lambda \sum_{c=1}^{C_{1:t}} -y^c(x_j;\theta_{1:t-1}^*) \log{y^c(x_j;\theta)}.
    \label{eq:lwf_loss}
\end{equation}
Similarly to \cref{eq:reg_loss}, $\theta$ is the weights of the main network that has been trained on a sequence of tasks so far and $\mathcal{L}_\text{t}(\theta)$ is cross-entropy loss on current classfication task $t$. $y(x_j;\theta_{1:t-1}^*)$ and $y(x_j;\theta)$ are the temperature-scaled logits of the old network and the current network, respectively. The main network that continuously learns data from task $1$ to $t-1$ is freezed and saved as the old network. $C_{1:t}$ denotes the total number of classes until task $t$ and thus, $y^c(x_j)$ refers to the $c^{th}$ output of the logit by an input $x_j$ from the current dataset $D_t$. Note that the second term in \cref{eq:lwf_loss} is also calculated on the current dataset as the previous data is not available. By emulating the soft targets generated by the previous model in addition to the ground truths in one-hot vector, the model can preserve the internal neural connection of the previous model while the weights are optimized for the new task. 

The logit scaled by temperature $\tau$ on the $c^{th}$ class position is calculated as follows:
\begin{equation}
    y^c(x_j;\theta) = \frac{(\boldsymbol{o}^c(x_j;\theta))^{1/\tau}}{\sum_k(\boldsymbol{o}^k(x_j;\theta))^{1/\tau}}, \qquad y^c(x_j;\theta_{1:t-1}^*) = \frac{(\boldsymbol{o}^c(x_j;\theta_{1:t-1}^*))^{1/\tau}}{\sum_k(\boldsymbol{o}^k(x_j;\theta_{1:t-1}^*))^{1/\tau}}
\end{equation}
where $\boldsymbol{o}(x_j;\theta)$ and $\boldsymbol{o}(x_j;\theta_{1:t-1}^*)$ are the outputs of the current network and the old network before softmax is applied. The higher temperature generates more evenly distributed soft targets. For example, if the temperature goes to infinity ($i.e.$ $\tau \to \infty$), $y(x_j)$ becomes an uniform vector ($i.e.$ $y^c(x_j) = 1/{C_{1:t}}$ for all $c\in{\{1, ..., C_{1:t}\} }$ ).

Another distillation method, \textit{less-forgetting learning} (LFL) \cite{jung2016less}, penalizes the differences of activations before last layer:
\begin{equation}
    \mathcal{L}_\text{LFL} = \mathcal{L}_\text{t}(\theta) + \lambda \norm{f(x_j;\theta) - f(x_j;\theta_{1:t-1}^*)}_2^2.
    \label{eq:lfl_loss}
\end{equation}
$f(x_j;\theta)$ and $f(x_j;\theta_{1:t-1}^*)$ are the centered and normalized activations of the main and old network respectively generated by input $x_j$ from the current dataset $D_t$ . The idea of LFL is the same as LwF except that the logits are replaced by the activations. The knowledge of the old tasks is retained by minimizing the gap between the activations from the previous model and the current model.

\subsection{Memory Replay Method}

Replay-based methods keep a part of the previous data (or exemplars) in a memory buffer. The memory buffer should contain the same number of exemplars for each class to build a balanced dataset and the few exemplars of each class should well represent the general features of their class. Then, a model is trained on the current dataset combined with the previous exemplars in the memory buffer to prevent the forgetting of the previous tasks. \emph{Incremental Classifier and Representation Learning} (iCaRL) \cite{rebuffi2017icarl} first proposes the usage of the memory buffer built on LwF \cite{li2017learning}. iCaRL calculates the mean of feature representations for each class and selects exemplars iteratively in a way that the mean of exemplars is closest to the class mean in feature representation space. This sampling strategy is called \emph{herding} and is well known to outperform a random sampling strategy at the cost of more computations. Moreover, iCaRL applies a nearest mean of exemplars classification instead of using a classifier layer, which classifies an image to the closest class mean of exemplar feature representations. Let $D_t$ be the current dataset of task $t$ and $P_{1:t-1}$ be the exemplar sets in the memory buffer which contains data from task $1$ to $t-1$. Then, the loss of iCaRL returns: 
\begin{equation}
    \mathcal{L}_\text{iCaRL} = \mathcal{L}_\text{t}(\theta) + \lambda \sum_{c=1}^{C_{1:t}} -y^c(x_j;\theta_{1:t-1}^*) \log{y^c(x_j;\theta)}
    \label{eq:icarl_loss}
\end{equation}
where the cross-entropy loss (the first term) and the distillation loss (the second term) are both calculated on the combined dataset $D_t^+ = D_t \cup P_{1:t-1}$. The rest of the notations are equal to \cref{eq:lwf_loss}.

\subsection{Bias Correction Method}
In memory replay methods, a network is trained on the combined dataset, a highly unbalanced dataset with few exemplars from the previous classes and sufficient data from the new classes. As a result, the network is biased towards the data of new classes which hold a large majority in the combined dataset. This problem is called \emph{task-recency bias} and recent articles show that alleviating the bias can significantly improve performance. 

For instance, \cite{wu2019large} proposes \emph{Bias Correction} (BiC) to prevent task-recency bias. They divide the combined dataset into a train set and a validation set and the model learns these sets continually through two-stage training. During the first stage, the model is trained on the train set with the loss in \cref{eq:icarl_loss}. In the second stage, they suggest the usage of a linear transformation on the logits $\boldsymbol{o}_k$ of new classes to compensate for the task-recency bias:
\begin{equation}
  q_k = \left \{
  \begin{aligned}
    &\boldsymbol{o}_k, && \text{if}\ 1 \le c \le C_{1:t-1}\\
    &\alpha \boldsymbol{o}_k + \beta, && \text{if}\ C_{1:t-1} < c \le C_{1:t}.
  \end{aligned} \right.
\end{equation} 
where the output of the old classes c ($1 \le c \le C_{1:t-1}$) remains the same and the output of the new classes c ($C_{1:t-1} < c \le C_{1:t}$) is linearly transformed by the learnable parameters $\alpha$ and $\beta$. These parameters are optimized by the validation set, while all other parameters in the network are frozen.  

Another bias correction method proposed by \cite{hou2019learning} is called \emph{Learning a Unified Classifier Incrementally via Rebalancing} (LUCIR) to tackle three problems that induce task-recency bias. The first problem they point out is that the feature norm of the new classes is larger than the feature norm of the old classes. In order to reduce the difference, they apply cosine normalization layer which are invariant to the magnitude of the feature instead of typical softmax layer. Then, the predicted probability $p_c(x)$ of class $c$ by input $x$ is calculated as follows:
\begin{equation}
    p_c(x) = \frac{\exp{(\theta_{L,c}^T f(x;\theta) +b_{L,c}})}{\sum_k \exp{(\theta_{L,k}^T f(x;\theta) +b_{L,k})}} \longrightarrow  p_c(x) =\frac{\exp{(\eta\langle\bar{\theta}_{L,c}, \bar{f}(x;\theta)\rangle})}{\sum_k \exp{(\eta\langle\bar{\theta}_{L,k}, \bar{f}(x;\theta)\rangle)}}
    \label{eq:applying_cosine_norm}
\end{equation}
where the left and right equations each represent the output probability of softmax layer and cosine normalization layer. $f$ is the feature extractor and $\theta_L$ and $b_L$ are the weights ($i.e.$ class embedding) and bias in the last layer $L$. $\bar{\theta}$ and $\bar{f}(x)$ denotes $l_2$ normlaized vector ($\bar{v} = v/\norm{v}_2$) and $\langle\bar{\theta}_{L,c}, \bar{f}(x)\rangle$ measures the cosine similarity between normalized weight and feature vector ($\langle\bar{v}_1, \bar{v}_2\rangle = \bar{v}_1^T\bar{v}_2$). The learnable scalar $\eta$ adjusts the peakness of softmax distribution since the range of $\langle\bar{\theta}_{L,c}, \bar{f}(x)\rangle$ is restricted to $[-1,1]$. 

The second problem is found in the distillation loss after applying cosine normalization in \cref{eq:applying_cosine_norm}: the angle between the feature vector $f(x;\theta)$ and the class embedding $\theta_{L,c}$ can be optimized to become similar as the angle between the previous feature vector $f(x;\theta_{1:t-1}^*)$ and the previous class embedding $\theta_{L,c}^*$ instead of being optimized to learn $f(x;\theta_{1:t-1}^*)$ itself. Therefore, the authors suggest the usage of cosine embedding loss which directly regularizes the angle between the feature vector $f(x;\theta)$ and the old feature vector $f(x;\theta_{1:t-1}^*)$:
\begin{equation}
    \mathcal{L}_\text{dis}(x) =\sum_{c=1}^{C_{1:t}} \norm{\langle\bar{\theta}_{L,c}, \bar{f}(x;\theta)\rangle- \langle\bar{\theta}^*_{L,c}, \bar{f}(x;\theta_{1:t-1}^*)\rangle}_2^2 \longrightarrow \mathcal{L}_\text{dis}(x) = 1-\langle\bar{f}(x;\theta), \bar{f}(x;\theta_{1:t-1}^*)\rangle.
\end{equation}
The equation on the left shows the previous distillation loss and the right one refers to the revised distillation loss.

The last problem addressed by \cite{hou2019learning} is inter-task confusion: the new class embeddings cluster together with the old class embeddings, which confuses the classification of the old and new classes. To prevent this, they employ \emph{margin ranking loss}:
\begin{equation}
    \mathcal{L}_\text{mr}(x) = \sum_{k=1}^K\max (m-\langle\bar{\theta}_L(x), \bar{f}(x;\theta)\rangle+\langle\bar{\theta}_L^k, \bar{f}(x;\theta)\rangle, 0 )
\end{equation}
where $\bar{\theta}_L(x)$ refers to the ground truth class embedding of $x$, $\theta_L^k$ is the embedding of top-$K$ closest classes, and $m$ is the margin threshold. This loss separate the current embeddings $\bar{\theta}_L(x)$ from the embeddings of K most similar class embedding $\bar{\theta}_L^k$. Combining all the solutions mentioned above, the loss of LUCIR finally returns:
\begin{equation}
    \mathcal{L}_\text{LUCIR} = \mathcal{L}_\text{t}(\theta) +  \lambda\mathcal{L}_\text{dis}(x_j) + \lambda_{mr} \mathcal{L}_\text{mr}(x_j)
    \label{eq:lucir_loss}
\end{equation}
where cross-entropy loss and distillation loss are measured by the combined dataset $D_t^+$ and a margin ranking loss is calculated on the previous exemplars $P_{1:t-1}$. $\lambda$ and $\lambda_{mr}$ are hyperparameters found by grid search.

Built upon LUCIR, recent work \cite{douillard2020podnet} suggests \emph{Pooled Outputs Distillation Network} (PODNet) which applies pooled out distillation loss and local similarity classifier. First, they define POD-width and POD-height losses as below:
\begin{align}
    \mathcal{L}_\text{POD-width}(f_l(x;\theta), f_l(x;\theta_{1:t-1}^*)) &= \sum_{c=1}^C \sum_{h=1}^H \left\lVert\sum_{w=1}^W f_{l,c,w,h}(x;\theta)-\sum_{w=1}^W f_{l,c,w,h}(x;\theta_{1:t-1}^*)\right\rVert_2^2\\
    \mathcal{L}_\text{POD-height}(f_l(x;\theta), f_l(x;\theta_{1:t-1}^*)) &= \sum_{c=1}^C \sum_{w=1}^W \left\lVert\sum_{h=1}^H f_{l,c,w,h}(x;\theta)-\sum_{h=1}^H f_{l,c,w,h}(x;\theta_{1:t-1}^*)\right\rVert_2^2
\end{align}
where $f_l(x;\theta)$ and $f_l(x;\theta_{1:t-1}^*)$ denotes the intermediate output of convolutional layer $l$ by a sample $x$ and both outputs are the representational matrix of size $C \times H \times W$. Thereafter, POD-width and POD-height losses are combined to build POD-spatial loss:
\begin{equation}
   \mathcal{L}_\text{POD-spatial}(f_l(x;\theta), f_l(x;\theta_{1:t-1}^*)) = \mathcal{L}_\text{POD-width}(f_l(x;\theta), f_l(x;\theta_{1:t-1}^*)) + \mathcal{L}_\text{POD-height}(f_l(x;\theta), f_l(x;\theta_{1:t-1}^*)). 
\end{equation}
POD-flat loss is further defined as $\mathcal{L}_\text{POD-flat}(f_L(x;\theta), f_L(x;\theta_{1:t-1}^*)) = \norm{f_L(x;\theta) -f_L(x;\theta_{1:t-1}^*)}_2^2$ with feature vector outputs of last convolutional layer $L$ in the main network and the old network. At last, all POD losses are combined to build POD-final loss:
\begin{align}
   \mathcal{L}_\text{POD-final}(x) = \lambda_c \sum_{l=1}^{L-1} \mathcal{L}_\text{POD-spatial}(f_l(x;\theta), f_l(x;\theta_{1:t-1}^*)) + \lambda_f \mathcal{L}_\text{POD-flat}(f_L(x;\theta), f_L(x;\theta_{1:t-1}^*))
\end{align}
where $\lambda_c$ and $\lambda_f$ adjusts the strength of two POD losses.

In addition, \cite{douillard2020podnet} indicates that the cosine normalization layer expressed in \cref{eq:applying_cosine_norm} optimizes a \emph{global similarity}: the training objective increases the cosine similarity between feature vector and weights pushing all feature vectors toward a single proxy (or mode) \cite{movshovitz2017no}. To improve it, they design Local Similarity Classifier (LSC) considering the usage of multiple proxies, which makes final embedding $f_L(x;\theta)$ robust to forgetting. In the setting of using $K$ proxies, the similarity $s_{c,k}$ of $k$th proxy for each class $c$ is first computed. Then, an averaged class similarity $y_c$ becomes the output of the classification layer:
\begin{align}
    s_{c,k}(x) = \frac{\exp\langle\theta_{L,c,k}, f_L(x;\theta)\rangle}{\sum_{i}\exp\langle\theta_{L,c,i}, f_L(x;\theta)\rangle}, \qquad y_c(x) = \sum_{k=1}^K s_{c,k}(x) \langle\theta_{L,c,k}, f_L(x;\theta)\rangle.
\end{align}
\cite{douillard2020podnet} empirically found that NCA loss \cite{goldberger2004neighbourhood} converges faster than cross-entropy loss, thereby modifying it to implement the LSC loss with a small margin $\delta$, a hinge $[\cdot]_+$, and a learnable parameter $\eta$:
\begin{align}
    \mathcal{L}_\text{LSC}(x) = \left[ -\log \frac{\exp (\eta(y_c(x;\theta)-\delta))}{\sum_{i \ne c} \exp \eta y_i(x;\theta)} \right]_+.
\end{align}
Finally, we obtain the loss of PODNet:
\begin{align}
    \mathcal{L}_\text{PODNET} =  \mathcal{L}_\text{LSC}(x_j) +  \mathcal{L}_\text{POD-final}(x_j)
    \label{eq:podnet_loss}
\end{align}
where the losses are calculated on the combined dataset $D_t^+$.

\section{The Application of ANCL to Methods in \cref{table:how_to_calculate_Omega} of Main Paper}\label{Appendix:C}

In this section, we explain the loss function of ANCL applied to methods in \cref{table:how_to_calculate_Omega}.
We first write down the loss of EWC and MAS on task $t$:
\begin{align}
    \mathcal{L}_\text{EWC} &= \mathcal{L}_\text{t}(\theta) + \frac{\lambda}{2}\sum_{i} F_{1:t-1,i}(\theta_{i}-\theta_{1:t-1,i}^{*})^2,
     \\
    \mathcal{L}_\text{MAS} &= \mathcal{L}_\text{t}(\theta) + \frac{\lambda}{2}\sum_{i} M_{1:t-1,i}(\theta_{i}-\theta_{1:t-1,i}^{*})^2.
     \label{eq:mas_loss}
\end{align}
The notation of above losses are the same as \cref{eq:reg_loss} except that $F_{1:t-1}$ is the diagonal elements of Fisher Information Matrix (FIM) that has been accumulated until task $t-1$. If $F_{1:t-1}$ is replaced with the importance $M_{1:t-1}$ defined by MAS \cite{aljundi2018memory}, it returns the loss of MAS in \cref{eq:mas_loss}. Then, ANCL can be applied to EWC~\cite{kirkpatrick2017overcoming} and MAS~\cite{aljundi2018memory} which generates \textit{Auxiliary Network EWC} (A-EWC) and \textit{Auxiliary Network MAS} (A-MAS) accordingly. The loss of A-EWC and A-MAS on task $t$ is defined as follows:
\begin{align}
    \mathcal{L}_\text{A-EWC} &= \mathcal{L}_\text{t}(\theta) + \frac{\lambda}{2}\sum_{i} F_{1:t-1,i}(\theta_{i}-\theta_{1:t-1,i}^{*})^2 + \frac{\lambda_a}{2}\sum_{i}F_{t,i}(\theta_{i}-\theta_{t,i}^{*})^2 \\  
    \mathcal{L}_\text{A-MAS} &= \mathcal{L}_\text{t}(\theta) + \frac{\lambda}{2}\sum_{i} M_{1:t-1,i}(\theta_{i}-\theta_{1:t-1,i}^{*})^2 + \frac{\lambda_a}{2}\sum_{i}M_{t,i}(\theta_{i}-\theta_{t,i}^{*})^2 \label{eq:a-mas_loss} 
\end{align}
where the importance $F_t$ and $M_t$ of the auxiliary parameters $\theta_{t}^{*}$ are calculated following the original methods (EWC or MAS) and $\lambda$ and $\lambda_a$ are fixed by grid search. The first two terms are equal to \cref{eq:reg_loss}.

Similarly, ANCL can be extended to the distillation-based methods such as \textit{Learning without Forgetting} (LwF)~\cite{li2017learning} in \cref{eq:lwf_loss} or \textit{less-forgetting learning} (LFL)~\cite{jung2016less} in \cref{eq:lfl_loss}. By applying ANCL to LwF and LFL, the new losses of \textit{Auxiliary Network LwF} (A-LwF) and \textit{Auxiliary Network LFL} (A-LFL) on task $t$ are written as follows:
\begin{align}
    \mathcal{L}_\text{A-LwF} &= \mathcal{L}_\text{t}(\theta) + \lambda  \sum_{c=1}^{C_{1:t}} -y^c(x_j;\theta_{1:t-1,i}^{*}) \log{y^c(x_j;\theta)} + \lambda_a \sum_{c=1}^{C_{1:t}} -y^c(x_j;\theta_t^*) \log{y^{c}(x_j;\theta)}, \label{eq:a-lwf_loss} \\
    \mathcal{L}_\text{A-LFL} &= \mathcal{L}_\text{t}(\theta) + \lambda \norm{f(x_j;\theta) - f(x_j;\theta_{1:t-1,i}^{*}) }_2^2 + \lambda_a \norm{f(x_j;\theta) - f(x_j;\theta_t^*)}_2^2, \label{eq:a-lfl_loss}
\end{align}
In \cref{eq:a-lwf_loss}, $y(x_j;\theta_t^*)$ represents the temperature-scaled logit of the auxiliary network and new regularization term is double summed over new class position $c$ and data $x_j$ of current task $t$. In \cref{eq:a-lfl_loss}, $f(x_j;\theta_t^*)$ is the normalized and centered activation of the auxiliary network. $\lambda$ and $\lambda_a$ are again found by grid search. 

In the same way, we can apply ANCL to memory replay and bias correction approaches (iCaRL \cite{rebuffi2017icarl}, BiC \cite{wu2019large}, LUCIR \cite{hou2019learning}, PODNet \cite{douillard2020podnet}). These methods usually use the memory buffer of the previous data and thus losses are calculated on the combined dataset $D_t^+$( the current dataset $D_t$ + the previous exemplars $P_{1:t-1}$).

First, we start from applying ANCL to the loss of iCaRL \cite{rebuffi2017icarl} in \cref{eq:icarl_loss}. Then, \emph{Auxiliary Network iCaRL} (A-iCaRL), A-iCaRL returns the loss function as below:
\begin{equation}
      \mathcal{L}_\text{A-iCaRL} = \mathcal{L}_\text{t}(\theta) + \lambda \sum_{c=1}^{C_{1:t}} -y^c(x_j;\theta_{1:t-1,i}^{*}) \log{y^c(x_j;\theta)} + \lambda_a \sum_{c=1}^{C_{1:t}} -y^c(x_j;\theta_t^*) \log{y^{c}(x_j;\theta)}
     \label{eq:a-icarl_loss}
\end{equation}
where the first two terms are the same as iCaRL loss and the last term represents a new regularizer based on the auxiliary network. $y^c(x_j;\theta_t^*)$ denotes the temperature-scaled logit of the auxiliary network.

Similarly, we apply ANCL to BiC \cite{wu2019large}, so called \emph{Auxiliary Network BiC} (A-BiC). A-BiC is also optimized by minimizing the loss function in \cref{eq:a-icarl_loss}. Note that BiC divides the combined dataset $D_t^+$ into a train set for main training stage and a validation set for bias correction stage. Two regularizers of A-BiC are measured on the train set like BiC.

Next, we adapt ANCL to the loss of LUCIR \cite{hou2019learning} in \cref{eq:lucir_loss} to build the loss of \emph{Auxiliary Network LUCIR} (A-LUCIR):
\begin{equation}
    \mathcal{L}_\text{A-LUCIR} = \mathcal{L}_\text{t}(\theta) +  \lambda \mathcal{L}_\text{dis}(x_j) + \lambda_a \mathcal{L}_\text{dis}^{aux}(x_j)+ \lambda_{mr} \mathcal{L}_\text{mr}(x_j)
\end{equation}
where $\mathcal{L}_\text{dis}^{aux}(x) = 1-\langle\bar{f}(x;\theta), \bar{f}(x;\theta_t^*)\rangle$ holds with the normalized feature vector $\bar{f}(x;\theta_t^*)$ of the auxiliary network.

Lastly, ANCL modifies PODNet loss (\cref{eq:podnet_loss}) to define \emph{Auxiliary Network PODNet} (A-PODNet) loss as follows:
\begin{equation}
    \mathcal{L}_\text{A-PODNET} =  \mathcal{L}_\text{LSC}(x_j) +  \lambda \mathcal{L}_\text{POD-final}(x_j) + \lambda_a \mathcal{L}_\text{POD-final}^{aux}(x_j)
\end{equation}
where $\mathcal{L}_\text{POD-final}^{aux}(x_j)$ stands for the POD loss with the feature representations of the auxiliary network as follows:
\begin{align}
   \mathcal{L}_\text{POD-final}^{aux}(x) = \lambda_c \sum_{l=1}^{L-1} \mathcal{L}_\text{POD-spatial}(f_l(x;\theta), f_l(x;\theta_t^*))) + \lambda_f \mathcal{L}_\text{POD-flat}(f_L(x;\theta)), f_L(x;\theta_t^*)))
\end{align}

Other than the above methods, ANCL can be adapted to most CL approaches with regularization terms based on the old parameters. However, it is hard to apply ANCL to the dynamic structure methods because these methods already contain an extra module or architecture to reflect plasticity.

\section{Comparison with AFEC}\label{Appendix:D}
In this section, we compare our ANCL with AFEC \cite{wang2021afec} which is the most similar method as ours. Before that, we formally introduce \textit{Active Forgetting with synaptic Expansion-Convergence} (AFEC) to clarify the difference easily. AFEC suggests to add an extra regularization term on EWC loss based on biologically inspired arguments. They argue that this term stimulates the active forgetting of previous knowledge that interferes with the new knowledge, whereas the existing regularization-based approaches highly concentrate on retaining the old weight. The loss of AFEC on task $t$ can be expressed as follows:
\begin{align}
    \mathcal{L}_\text{AFEC} &= \mathcal{L}_\text{t}(\theta) + \frac{\lambda}{2}\sum_{i} F_{1:t-1,i}(\theta_{i}-\theta_{1:t-1,i}^{*})^2 + \frac{\lambda_a}{2}\sum_{i} F_{t,i}(\theta_{i}-\hat{\theta}_{t,i})^2. \label{eqn:AFEC_loss}
\end{align}
The first two terms are the same as \cref{eq:ewc_loss}, while the last promotes active forgetting by regularizing the parameters $\theta\in\mathbb{R}^P$ towards the biologically inspired \textit{expanded parameters} $\hat{\theta}_t\in\mathbb{R}^P$~(see precise definition in~\cite{wang2021afec}) through FIM $F_t$ on task $t$. The expanded parameters are solely trained on the current dataset allowing the forgetting of the old datasets and $\lambda_a$ is a hyperparameter determined by grid search. As a result, the main network parameters $\theta$ efficiently learns from both the old parameters $\theta_{1:t-1}^{*}$ and the expanded parameters $\hat{\theta}_{t}$. Moreover, the last term of \cref{eqn:AFEC_loss} can be applied to other regularization-based methods as a \textit{plug-and-play} method. Thus, A-EWC is equivalent to AFEC\footnote{However, A-MAS is not equal to AFEC as the new regularizer of ANCL is based on MAS importance not FIM.} except that the importance $F_t$ of the auxiliary network in ANCL is calculated only once before the training of the new task and then fixed afterward.\footnote{The code implementation of AFEC newly calculates $F_t$ every epoch.} 

The expanded parameters $\hat\theta_t$ in AFEC are computed as follows: at the beginning of each task, the expanded parameters are initialized by the old parameters, Then, at each epoch, it is trained on task $t$ without regularization, returning the network parameters $\hat\theta_t$. For the current epoch of the original network, these weights are then taken into account to modify the regularized dynamics. This procedure is repeated every epoch, and hence for each epoch the dynamics of SGD is augmented with a freshly computed regularizer. 

Finally, we evaluate ANCL and AFEC in \cref{table:AFEC_ANCL_compare} and \cref{table:AFEC_ANCL_compare_replay}. \cref{table:AFEC_ANCL_compare} demonstrates that ANCL outperforms AFEC in every method on task incremental scenario. Specifically, AFEC achieves similar improvement as ANCL for EWC while obtaining very marginal improvement for other methods. This results agree with our description above that EWC with AFEC and EWC with ANCL have the equivalent loss and the difference comes from how the new regularizer is calculated. Except EWC, ANCL can balance two regularizers better than AFEC which results in higher improvement in accuracy. In \cref{table:AFEC_ANCL_compare_replay}, we also compare AFEC and ANCL on class incremental scenario. Although AFEC gives some improvements, ANCL outperforms AFEC on all method. These results support that ANCL can effectively utilize the auxiliary network compared to AFEC which applies the fixed regularizer to all methods. In conclusion, the advantage of ANCL lies on its natural adaptation of the regularizer following the original methods such that the optimal solution well associate the old network with the auxiliary network. 

\begin{table}
\centering
\begin{tabular}{cccc}
\hline
Methods & CL (original)    & w/ AFEC \cite{wang2021afec} & w/ ANCL (ours) \\ \hline
EWC \cite{kirkpatrick2017overcoming}     & $58.13_{\pm0.87}$  & $60.60_{\pm1.63}$   & $60.86_{\pm1.46}$  \\
MAS \cite{aljundi2018memory}             & $60.56_{\pm0.82}$  & $61.28_{\pm0.81}$   & $64.43_{\pm1.17}$   \\
LwF \cite{li2017learning}                & $78.87_{\pm0.69}$  & $78.77_{\pm0.72}$   & $79.42_{\pm0.57}$   \\
LFL \cite{jung2016less}                  & $74.50_{\pm0.57}$  & $74.55_{\pm0.62}$   & $75.23_{\pm0.67}$   \\ \hline
\end{tabular}
\caption{The averaged accuracy (\%) on benchmark (1) CIFAR-100/10. Reported metrics are averaged over 3 runs (averaged accuracy $\pm$ standard error). AFEC \cite{wang2021afec} and ANCL (ours) are applied to CL approaches and compared.} 
\label{table:AFEC_ANCL_compare}
\end{table}

\begin{table}
\centering
\begin{tabular}{cccc}
\hline
Methods & CL (original) & w/ AFEC \cite{wang2021afec} & w/ ANCL (ours) \\ \hline
iCaRL \cite{rebuffi2017icarl}        & $58.05_{\pm0.94}$ & $59.31_{\pm0.97}$   & $61.22_{\pm0.88}$   \\
BiC \cite{wu2019large}               & $56.74_{\pm1.33}$ & $57.08_{\pm1.22}$   & $58.32_{\pm1.27}$   \\
LUCIR \cite{hou2019learning}         & $56.06_{\pm0.45}$ & $58.97_{\pm0.92}$   & $60.20_{\pm0.78}$   \\
PODNet \cite{douillard2020podnet}    & $61.80_{\pm0.77}$ & $62.73_{\pm0.68}$   & $63.15_{\pm0.62}$   \\ \hline
\end{tabular}
\caption{The averaged incremental accuracy (\%) on benchmark (5) CIFAR-100/6. Reported metrics are averaged over 3 runs (averaged accuracy $\pm$ standard error). AFEC \cite{wang2021afec} and ANCL (ours) are applied to CL approaches and compared.} \label{table:AFEC_ANCL_compare_replay}
\end{table}

\section{The Mathematical Analysis of ANCL Gradients}\label{Appendix:E}

In this section, we closely look into CL and ANCL by analyzing their gradients. For EWC~\cite{kirkpatrick2017overcoming} and MAS~\cite{aljundi2018memory}, we directly solve the optimal weights of CL and ANCL from their gradients and analyze them in respect of the stability-plasticity balance. For LwF~\cite{li2017learning}, and LFL~\cite{jung2016less}, we compare the gradient of CL and ANCL after some approximations. Then, we explain that other methods with memory buffer (iCaRL~\cite{rebuffi2017icarl}, BiC~\cite{wu2019large}, LUCIR~\cite{hou2019learning}, and PODNet~\cite{douillard2020podnet}) are originated from distillation-based method (LwF and LFL). Therefore, previous analysis on LwF and LFL can be extended and applied to their variations respectively.   

First, we start with the EWC loss in \cref{eq:ewc_loss}. The gradient with respect to the $i^{th}$ parameter $\theta_i$ of the model weights $\theta = (\theta_1,\dots,\theta_P)\in\mathbb{R}^P$ reads:
\begin{equation}
    \nabla_{\theta_i} \mathcal{L}_\text{EWC} = \nabla_{\theta_i} \mathcal{L}_\text{t}(\theta) + \lambda F_{1:t-1,i} (\theta_i-\theta_{1:t-1,i}^{*}) .
    \label{eq:grad_ewc_loss}
\end{equation}
It consists of two terms: the first term updates $\theta_i$ toward the minima of the task-specific loss of the current task and the second term regularizes the model weight $\theta_i$ to be as close as possible to the old weight $\theta_{1:t-1,i}^{*}$ according to $\lambda F_{1:t-1,i}$. Then, the $i^{th}$ model parameter is updated from the $k^{th}$ to the $(k+1)^{th}$ iteration with learning rate $\eta$ as follows:
\begin{equation}
    \theta_i^{(k+1)} \leftarrow \theta_i^{(k)}  - \eta \nabla_{\theta_i^{(k)}} \mathcal{L}_\text{EWC}
    \label{eq:ewc_update}
\end{equation}
According to \cite{lubana2021quadratic}, the weight of the $k^{th}$ iteration $\theta_i^{(k)}$ can be expressed by the initial weight $\theta_{1:t-1,i}^{*}$ via the recursive substitution of \cref{eq:ewc_update} during the $k^{th}$ iteration: 
\begin{equation}
    \theta_i^{(k)} = \theta_{1:t-1,i}^{*} - \sum_{l=0}^{k-1} [(1-\eta\lambda F_{1:t-1,i})^{(k-l-1)}\eta] g_i^{(l)} ,
    \label{eq:ewc_kth_iter}
\end{equation}
where $g_i^{(l)}$ denotes the gradient of task specific loss with respect to $\theta_i^{(l)}$ at the $l^{th}$ iteration. If we take $k \to \infty$ on the both side of \cref{eq:ewc_kth_iter}, we can obtain the optimal parameter $\theta_{1:t,i}^{*}= \lim_{k \to \infty} \theta_i^{(k)} $ on task $t$ :
\begin{equation}
    \theta_{1:t,i}^{*} = \underbrace{\theta_{1:t-1,i}^{*}}_\text{Previous param.}
    - \underbrace{\lim_{k \to \infty} \sum_{l=0}^{k-1} [(1-\eta\lambda F_{1:t-1,i})^{(k-l-1)}\eta] g_i^{(l)}}_\text{Task-specific updates}.
    \label{eq:ewc_optimal_solution}
\end{equation}
\cref{eq:ewc_optimal_solution} shows that the optimal weight $\theta_{1:t,i}^{*}$ after task $t$ is obtained from the previous parameter $\theta_{1:t-1,i}^{*}$ and the sequence of task-specific updates $g_i^{(l)}$ which is adjusted by the effective learning rate $(1-\eta\lambda F_{1:t-1,i})^{(k-l-1)}\eta$. Since the learning rate $\eta (>0)$ and the importance of each parameter $F_{1:t-1,i} (>0)$ are given from the beginning, the effective learning rate is fully depends on hyperparameter $\lambda$. For high $\lambda$, the weight updates are restricted as the value of $1-\eta\lambda F_{1:t-1,i}$ becomes smaller which consequently reduces the effective learning rate. For low $\lambda$, the effective learning rate becomes approximately equal to $\eta$, which allows the free update of the weight on a new task. It is also empirically shown in \cite{lubana2021quadratic} that when $(1-\eta\lambda F_{1:t-1,i})>1$ or $(1-\eta\lambda F_{1:t-1,i})<0$ holds, the training might become unstable because the effective learning rate will grow exponentially or change its sign every update.  

Next, we analyze A-EWC loss in the same way. The gradient of \cref{eq:a-ewc_loss} on $i^{th}$ parameter $\theta_i$ shows:
\begin{equation}
    \nabla_{\theta_i} \mathcal{L}_\text{A-EWC} = \nabla_{\theta_i} \mathcal{L}_\text{EWC} + \lambda_a F_{t,i}(\theta_i-\theta_{t,i}^{*}).
    \label{eq:grad_aewc_loss}
\end{equation}
Similarly to EWC, $\theta_i$ of A-EWC is updated from the $k^{th}$ to the $(k+1)^{th}$ iteration with learning rate $\eta$ as follows:
\begin{equation}
    \theta_i^{(k+1)} \leftarrow \theta_i^{(k)}  - \eta \nabla_{\theta_i^{(k)}} \mathcal{L}_\text{A-EWC}
    \label{eq:aewc_update}
\end{equation}
Then, we can extend the derivation of \cref{eq:ewc_kth_iter} following \cite{lubana2021quadratic} to A-EWC through the recursive substitution of \cref{eq:aewc_update}. Then, the updated weight $\theta_i^{(k)}$ at $k^{th}$ iteration initialized with the previous optimal weight $\theta_{1:t-1,i}^{*}$ returns:
\begin{equation}
    \theta_i^{(k)} = (1-\alpha-\beta)^k \theta_{1:t-1,i}^{*} +\sum_{l=0}^{k-1}(1-\alpha-\beta)^l (\alpha\theta_{1:t-1,i}^{*} + \beta\theta_{t,i}^{*} )  - \sum_{l=0}^{k-1} [(1-\alpha-\beta)^{(k-l-1)}\eta] g_i^{(l)}
    \label{eq:aewc_kth_iter}
\end{equation}
where $\alpha = \eta\lambda F_{1:t-1,i}$ and $\beta = \eta\lambda_a F_{t,i}$ hold. The detailed derivation of \cref{eq:aewc_kth_iter} can be found in \cref{appendix:iterative_formula_proof}.

Based on the experiment results of \cite{lubana2021quadratic}, one can similarly argue that $0\le(1-\alpha-\beta)<1$ is a sufficient condition for stable training with \cref{eq:aewc_kth_iter}. Thus, we can safely assume that $\lambda>0$ and $\lambda_a>0$ are chosen appropriately where $(1-\alpha-\beta)<0$ is not a case.  Based on the assumption, we take $k \to \infty$ on \cref{eq:aewc_kth_iter}. Then, the first term $(1-\alpha-\beta)^k \theta_{1:t-1,i}^{*}$ converges to zero as $\lim_{k \to \infty} (1-\alpha-\beta)^k = 0 $ and the second term can be further calculated using the infinite sum of geometric sequence ($i.e.$ $\sum_{n=0}^{\infty} ar^n = a/(1-r)$ for $|r|<1$ and $a \ne 0$). Finally, the optimal parameter $\theta_{1:t,i}^* = \lim_{k \to \infty} \theta_i^{(k)}$ can be simply expressed as:
\begin{equation}
\theta_{1:t,i}^{*} =  \underbrace{\frac{\alpha\theta_{1:t-1,i}^{*} + \beta\theta_{t,i}^{*}}{\alpha + \beta}}_\text{Interpolated param.} - \underbrace{\lim_{k \to \infty} \sum_{l=0}^{k-1} [(1-\alpha-\beta)^{(k-l-1)}\eta] g_i^{(l)}}_\text{Task-specific updates}.
\label{eq:aewc_optimal_solution}
\end{equation}
In \cref{eq:aewc_optimal_solution}, the optimal weight $\theta_{1:t,i}^{*}$ after task $t$ is determined by the interpolation of the old parameter $\theta_{1:t-1,i}^{*}$ and the auxiliary parameter $\theta_{t,i}^{*}$ and the sequence of task-specific updates $g_i^{(l)}$ with the effective learning rate $(1-\alpha-\beta)^{(k-l-1)}\eta$. First, the interpolated parameter is determined by the relative ratio between $\alpha$ and $\beta$. If $\alpha > \beta$ is the case, the interpolation will lean toward the old parameter and if $\alpha < \beta$ holds, the interpolation will be located closer to the auxiliary parameter. Moreover, the effective learning rate now depends on both $\lambda$ and $\lambda_a$. Thus, if $\lambda$ and $\lambda_a$ are both high, the effective learning rate will become smaller and consequently less task-specific updates will be made. As a result, the optimal weight will converge to the interpolation of two network parameters. Otherwise, the parameter will be simply optimized for the current task via task-specific updates. Compared to \cref{eq:ewc_optimal_solution} where high $\lambda$ interferes with learning a new task, the old parameter with high $\lambda$ can still be updated toward the auxiliary parameter through the interpolation. 

For MAS and A-MAS loss (\cref{eq:mas_loss} and \cref{eq:a-mas_loss}), a similar analaysis as EWC and A-EWC can be applied if we simply replace the regularization factor $F$ of EWC with the regularization factor $M$ of MAS. In fact, the solutions of ANCL for EWC and MAS are located on the interpolation between the old weights and the auxiliary weights and reaches the highest CKA scores with the multitask model on it, which can be seen in the trade-off analysis (\cref{sec:trade-off_analysis}) of the main paper.

Among distillation losses, we first take a derivative on LFL loss (\cref{eq:lfl_loss}) with respect to the model parameters $\theta$:
\begin{align}
    \nabla_{\theta} \mathcal{L}_\text{LFL} &= \underbrace{\nabla_{\theta} \mathcal{L}_\text{t}(\theta)}_\text{Task-specific grad.} + \underbrace{2 \lambda (f(x_j;\theta) - f(x_j;\theta_{1:t-1}^*)) \nabla_{\theta} f(x_j;\theta)}_\text{Distillation grad.}.
\end{align}
The gradient of LFL loss can be divided into two part parts: the task-specific gradient that drives $\theta$ toward the minima of current task loss and the distillation gradient that transfers the knowledge of old network to the current network by minimizing the difference of activations. Compared to EWC and MAS, LFL has more flexibility to memorize the previous knowledge since it doesn't directly regularize its weight but activations. The model trained with too high $\lambda$ will mainly focus on generating exactly the same activations as the old model while neglecting learning a new task.

Next, we take the same derivative on A-LFL loss in \cref{eq:a-lfl_loss}: 
\begin{equation}
      \nabla_{\theta} \mathcal{L}_\text{A-LFL} = \nabla_{\theta} \mathcal{L}_\text{t}(\theta) + 2 \lambda (f(x_j;\theta) - f(x_j;\theta_{1:t-1}^*)) \nabla_{\theta} f(x_j;\theta) +  2 \lambda_a (f(x_j;\theta) - f(x_j;\theta_{t}^*)) \nabla_{\theta} f(x_j;\theta)   
\end{equation}
which can be organized as: 
\begin{equation}
    \nabla_{\theta} \mathcal{L}_\text{A-LFL} = \underbrace{\nabla_{\theta} \mathcal{L}_\text{t}(\theta)}_\text{Task-specific grad.} + \underbrace{2 (\lambda + \lambda_a) (f(x_j;\theta) -  \frac{\lambda f(x_j;\theta_{1:t-1}^*) + \lambda_a f(x_j;\theta_{t}^*)}{\lambda + \lambda_a}  ) \nabla_{\theta} f(x_j;\theta)}_\text{Double distillation grad.}.
    \label{eq:grad_alfl_loss}
\end{equation}
Then, the double distillation gradient moves the old and new knowledge from the two networks (the old and auxiliary networks) into the current network by driving the activation of the main network to the interpolated activation between the auxiliary network and the old network. The interpolation is decided by the ratio of $\lambda$ and $\lambda_a$ which strike a balance between stability (old network) and plasticity (auxiliary network). Moreover, it is worth to note that \cref{eq:grad_alfl_loss} becomes a similar form as the graident of A-EWC if we assume one-hidden-layer network where $f(x_j;\theta_{1:t-1}^*)=W_{old} x_j$, $f(x_j;\theta_{t}^*)=W_{aux} x_j$, and $f(x_j;\theta)=W x_j$ hold for $W_{old},W_{aux},W \in \mathbb{R}^{p \times d}$ and $x_j \in \mathbb{R}^{d\times 1}$:
\begin{equation}
    \nabla_{\theta} \mathcal{L}_\text{A-LFL} = \nabla_{\theta} \mathcal{L}_\text{t}(\theta) + 2 (\lambda + \lambda_a) (W -  \frac{\lambda W_{old} + \lambda_a W_{aux}}{\lambda + \lambda_a}  ) x_j x_j^T .
\end{equation}
Now, the second term drifts the model weight $W$ (except last layer) toward the interpolation of the old weight $W_{old}$ and the auxiliary weight $W_{aux}$ like the gradient of A-EWC. 

Next, let's consider the loss of LwF (\cref{eq:lwf_loss}). For simplicity, we assume that $y^c(x_j;\theta)$ and $y^c(x_j;\theta_{1:t-1}^*)$ are softmax logits scaled by temperature $\tau$ such as:
\begin{align}
    y^c(x_j;\theta) = \frac{e^{\boldsymbol{o}^c(x_j;\theta)/\tau}}{\sum_k e^{\boldsymbol{o}^k(x_j;\theta)/\tau}}, \qquad y^c(x_j;\theta_{1:t-1}^*) = \frac{e^{\boldsymbol{o}^c(x_j;\theta_{1:t-1}^*)/\tau}}{\sum_k e^{\boldsymbol{o}^k(x_j;\theta_{1:t-1}^*)/\tau}}.
    \label{eq:softmax}
\end{align}
Then, referring to \cite{hinton2015distilling}, the gradient of LwF with respect to $\theta$ can be calculated as follows (detailed proof in \cref{appendix:hinton_proof}):
\begin{align}
    \nabla_{\theta} \mathcal{L}_\text{LwF} &= \nabla_{\theta} \mathcal{L}_\text{t}(\theta) + \frac{\lambda}{\tau} (y(x_j;\theta) -  y(x_j;\theta_{1:t-1}^*)) \nabla_{\theta} \boldsymbol{o}(x_j;\theta).
    \label{eq:hinton_KD}
\end{align}
When $\tau$ is sufficiently large, the softmax in \cref{eq:softmax} can be approximated using $exp(\boldsymbol{o}(x_j;\theta)/\tau) \approx 1 + \boldsymbol{o}(x_j;\theta)/\tau$ which substitute $y(x_j;\theta)$ and $y(x_j;\theta_{1:t-1}^*)$ in \cref{eq:hinton_KD}:
\begin{equation}
    \nabla_{\theta} \mathcal{L}_\text{LwF} \approx \nabla_{\theta} \mathcal{L}_\text{t}(\theta) + \frac{\lambda}{\tau}(\frac{1+\boldsymbol{o}(x_j;\theta)/\tau}{C_{1:t} + \sum_k \boldsymbol{o}^k(x_j;\theta)/\tau} - \frac{1+\boldsymbol{o}(x_j;\theta_{1:t-1}^*)/\tau}{C_{1:t} + \sum_k \boldsymbol{o}^k(x_j;\theta_{1:t-1}^*)/\tau}) \nabla_{\theta} \boldsymbol{o}(x_j;\theta)
\end{equation}
We further assume the logits of the main network and the old network have zero-mean ($i.e.$ $\sum_k \boldsymbol{o}^k(x_j;\theta)=0$ and $\sum_k \boldsymbol{o}^k(x_j;\theta_{1:t-1}^*)=0$). Finally, we obtain the following approximation:
\begin{align}
    \nabla_{\theta} \mathcal{L}_\text{LwF} \approx \underbrace{\nabla_{\theta} \mathcal{L}_\text{t}(\theta)}_\text{Task-specific grad.} + \underbrace{\frac{\lambda}{C_{1:t}\tau^2}  (\boldsymbol{o}(x_j;\theta) - \boldsymbol{o}(x_j;\theta_{1:t-1}^*)) \nabla_{\theta} \boldsymbol{o}(x_j;\theta)}_\text{Distillation grad.}.
\end{align}
This demonstrates that the gradient of LwF is in the equivalent form as the gradient of LFL under sufficiently large temperature and zero-mean logits of both main and old network. Thereby, similar interpretation as LFL is applicable. One should just note that $\boldsymbol{o}(x_j;\theta)$ in LwF is a logit and $f(x_j;\theta)$ in LFL is an activation before last layer. 

Similarly, we can approximate the gradient of A-LwF in \cref{eq:a-lwf_loss}:
\begin{multline}
    \nabla_{\theta} \mathcal{L}_\text{A-LwF} \approx \nabla_{\theta} \mathcal{L}_\text{t}(\theta) + \frac{\lambda}{C_{1:t}\tau^2} (\boldsymbol{o}(x_j;\theta) - \boldsymbol{o}(x_j;\theta_{1:t-1}^*)) \nabla_{\theta} \boldsymbol{o}(x_j;\theta) \\+ \frac{\lambda_a}{C_{1:t}\tau^2} (\boldsymbol{o}(x_j;\theta) - \boldsymbol{o}(x_j;\theta_{t}^*)) \nabla_{\theta} \boldsymbol{o}(x_j;\theta).
\end{multline}
Then, we get the final form as follows:
\begin{equation}
    \nabla_{\theta} \mathcal{L}_\text{A-LwF} \approx \underbrace{\nabla_{\theta} \mathcal{L}_\text{t}(\theta)}_\text{Task-specific grad.} + \underbrace{\frac{\lambda+\lambda_a}{C_{1:t}\tau^2}(\boldsymbol{o}(x_j;\theta) - \frac{\lambda \boldsymbol{o}(x_j;\theta_{1:t-1}^*)+\lambda_a \boldsymbol{o}(x_j;\theta_{t}^*)}{\lambda+\lambda_a} ) \nabla_{\theta} \boldsymbol{o}(x_j;\theta)}_\text{Double distillation grad.}.
    \label{eq:grad_alwf_loss}
\end{equation}
Similar to A-LFL, the second part of gradient shift the logit of the main network toward the interpolation of the logits from the old network and the new network. The stability-plasticity dilemma is again solved by directly controlling $\lambda$ and $\lambda_a$.

Unlike the analysis of EWC and MAS, we cannot claim that the optimal weight of A-LFL and A-LwF will be located on the interpolation between the old and auxiliary weights (actually, it is not true referring to \cref{fig:trade-off_figures} in the main paper). Instead, we can at least claim that the double distillation gradient added to the task-specific gradient in \cref{eq:grad_alfl_loss} and \cref{eq:grad_alwf_loss} tends to derive the activation of main network to the interpolated activation between the auxiliary and old network. In the trade-off analysis (\cref{sec:trade-off_analysis}) of the main paper, it is confirmed that the solutions of LwF and LFL tends to move toward the weight space between the old weights and the auxiliary weights and the highest CKA score with the multitask model is also achieved in the middle.

Other four methods (iCaRL~\cite{rebuffi2017icarl}, BiC~\cite{wu2019large}, LUCIR~\cite{hou2019learning}, and PODNet~\cite{douillard2020podnet}) for class incremental learning are basically originated from distillation-based methods (LwF and LFL). Thus, our previous analyses can be extended or reused according to their variations. For example, iCaRL and BiC are both based on the LwF loss (see \cref{eq:icarl_loss}), except the facts that both methods retain a memory buffer and BiC has additional bias correction stage. Therefore, the analysis of LwF still holds for these two methods. LUCIR (\cref{eq:lucir_loss}) also adopts the distillation loss with the margin ranking loss which is aligned with the extension of distillation-based methods. In PODNet, pooled outputs distillation (POD) loss (\cref{eq:podnet_loss}) constrains the output of each intermediate convolutional layer, which is similar to LFL loss that restricts only the final output. Thus, extending the analysis of LFL to all outputs of intermediate layers will be sufficient to explain the gradient of PODNet. 

In conclusion, CL losses bind the training dynamic of the current model with corresponding old one through hyperparameter $\lambda$. Thus, the high value of $\lambda$ hinders learning a new data as it reduces the effective learning rate of the task-specific gradient or increases the size of the distillation gradient. However, ANCL loss function promotes the learning of a new data through the interpolated parameters or the interpolated activations in the double distillation gradient even when the current model is highly regularized toward the old model. This interpolated region is turned out to be a better trade-off between stability and plasticity by \cref{sec:trade-off_analysis} in the main text.

\section{Details on Experiments}\label{Appendix:F}

\subsection{Implementation Details}\label{Appendix:F.1}
The model is trained from scratch and every experiment is conducted 3 times with different seeds to generate averaged metrics. The class order is fixed following iCaRL \cite{rebuffi2017icarl} to reduce variance in results such that each task always consists of the same set of classes. We build our code implementation based on Framework for Analysis of Class-Incremental Learning (FACIL) \cite{masana2020class} which supports several benchmarks and implements existing CL methods. SGD optimizer with momentum 0.9 and batch size 128 is applied to all experiments. The initial learning rate of each task (for task incremental learning) or phase (for class incremental learning) is chosen among the set of learning rate {[}0.1, 0.05, 0.01, 0.005, 0.001{]} by grid search. The learning rate of the first task or initial phase is always chosen slightly bigger than the following tasks since the model is trained from scratch. Furthermore, the learning rate is decreased by the factor of 3 whenever there is no improvement in validation loss during 5 epochs. When the learning rate reaches a given minimum threshold (0.0001), training is stopped. The model can be trained for a maximum of 200 epochs per task. Lastly, we conduct all experiments using gpu ”NVIDIA GeForce GTX 1080 Ti”.

\subsection{Detail Results on Task Incremental Scenario}\label{Appendix:F.2}
\begin{figure}[htbp]
\centerline{\includegraphics[scale=.55]{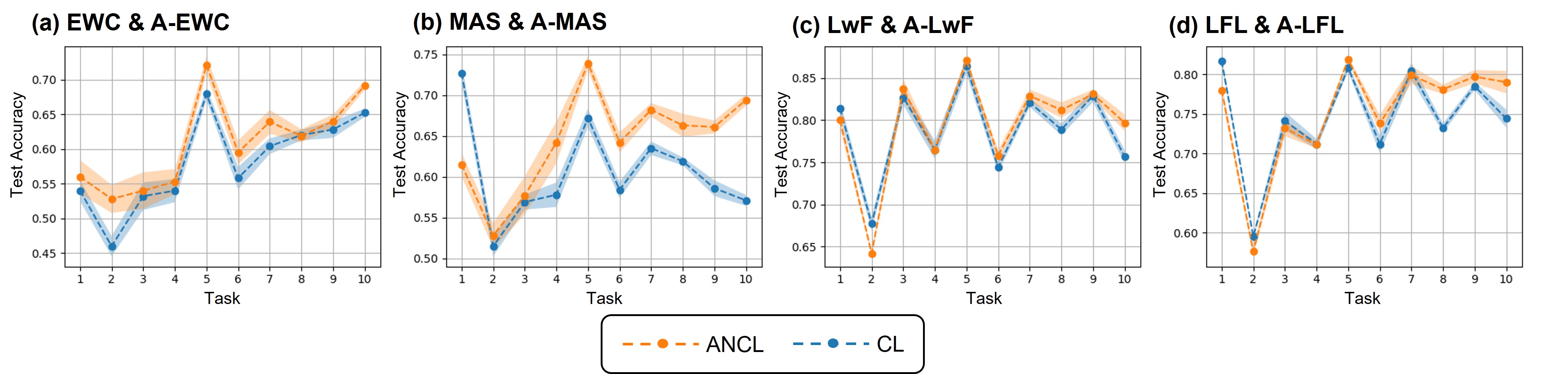}}
\caption{The final accuracy on all tasks of (1) CIFAR-100/10 with its standard error as an error band. Orange line represents ANCL methods ((a) A-EWC, (b) A-MAS, (c) A-LwF, and (d) A-LFL and blue line visualizes CL methods ((a) EWC, (b) MAS, (c) LwF, and (d) LFL).}
\label{fig:TIL_final_accuracy}
\end{figure}
In addition to the averaged accuracy table in the main paper, we plot the final accuracy of the sequential task on benchmark (1) CIFAR-100/10 in \cref{fig:TIL_final_accuracy}. In every figure, the accuracy of most tasks of ANCL approaches are higher than that of CL approaches. Concretely, ANCL achieves better performance in the later task compare to CL at the cost of losing earlier task accuracy, which is well shown in Figure  (b), (c), and (d). This is because ANCL methods is able to learn a new task better than CL through better stability-plasticity balance. As a trade-off, ANCL losses a bit of ability to remember initial knowledge, but ANCL is still comparable to CL in earlier tasks. 

\subsection{Detail Results on Class Incremental Scenario}\label{Appendix:F.3}
\begin{figure}[htbp]
\centerline{\includegraphics[scale=.55]{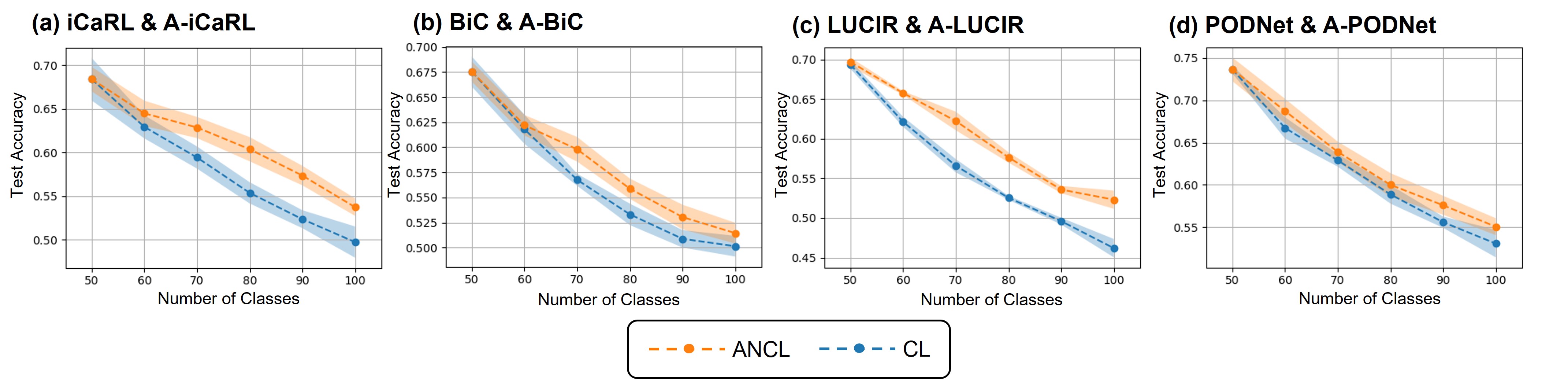}}
\caption{The accuracy at each phase on (5) CIFAR-100/6 with its standard error as an error band. Orange line represents ANCL methods ((a) A-iCaRL, (b) A-BiC, (c) A-LUCIR, and (d) A-PODNet) and blue line visualizes CL methods ((a) iCaRL, (b) BiC, (c) LUCIR, and (d) PODNet).}
\label{fig:CIL_final_accuracy}
\end{figure}

\begin{table*}[htbp]
\centering
\begin{tabular}{ccccc}
\hline
                              & \multicolumn{2}{c}{CIFAR-100}          & \multicolumn{2}{c}{Tiny ImageNet} \\
Methods                       & (5)                  & (6)                     & (7)                   & (8)               \\ \hline
Fine-tuning                   & $40.63_{\pm0.93}$    & $38.09_{\pm0.42}$    & $18.80_{\pm0.55}$     & $16.34_{\pm0.44}$     \\
Joint                         & $66.10_{\pm1.04}$    & $64.40_{\pm0.83}$    & $44.01_{\pm0.34}$     & $43.54_{\pm0.37}$     \\
EEIL \cite{castro2018end}     & $41.61_{\pm1.10}$     & $39.48_{\pm0.27}$     & $20.99_{\pm0.24}$     & $19.26_{\pm0.96}$ \\ \hline
iCaRL \cite{rebuffi2017icarl} & $49.75_{\pm0.93}$         & $45.85_{\pm0.89}$        & $27.83_{\pm0.43}$             & $29.26_{\pm0.86}$          \\
\rowcolor{Gray}
w/ ANCL (ours)             & $53.87_{\pm0.59}$         & $48.72_{\pm0.33}$        & $30.36_{\pm0.97}$             & $31.32_{\pm0.41}$        \\
BiC \cite{wu2019large}      & $50.13_{\pm0.77}$         & $47.97_{\pm1.17}$        & $32.87_{\pm0.19}$             & $29.45_{\pm0.54}$      \\
\rowcolor{Gray}
w/ ANCL (ours)               & $51.35_{\pm0.89}$         & $50.44_{\pm1.31}$        & $34.26_{\pm0.34}$             & $32.40_{\pm0.39}$        \\ \hline
LUCIR \cite{hou2019learning}     & $46.19_{\pm0.52}$         & $45.89_{\pm0.68}$        & $24.07_{\pm0.46}$             & $20.89_{\pm0.78}$         \\
\rowcolor{Gray}
w/ ANCL (ours)                    & $52.30_{\pm0.48}$         & $49.07_{\pm0.43}$        & $28.96_{\pm0.58}$             & $23.65_{\pm0.55}$         \\ \hline
PODNet \cite{douillard2020podnet} & $53.05_{\pm0.89}$         & $49.63_{\pm0.83}$        & $30.78_{\pm0.27}$             & $30.25_{\pm0.52}$            \\
\rowcolor{Gray}
w/ ANCL (ours)                    & $55.01_{\pm0.32}$         & $51.39_{\pm0.62}$        & $32.53_{\pm0.50}$             & $33.11_{\pm0.61}$           \\ \hline
\end{tabular}
\caption{The final accuracy (\%) on benchmark (5)-(8). Reported metrics are averaged over 3 runs (averaged accuracy $\pm$ standard error). ANCL methods are colored in gray.} 
\label{table:CIL_final_accuracy}
\end{table*}

Similar to \cref{fig:TIL_final_accuracy}, we plot the accuracy at each phase on class incremental scenario in \cref{fig:CIL_final_accuracy}. ANCL and CL are trained on initial 50 classes and then incrementally learn 10 classes per phase on CIFAR-100. In all methods, ANCL outperforms CL in every phase. Especially, LUCIR shows the biggest improvement and PODNet shows the smallest improvement among 4 methods. Even though ANCL naturally mimicks the regularizer of the original method to adopt plasticity in their method, some methods are less compatible with ANCL.

The final accuracy of Tab. 3 in the main paper is shown in \cref{table:CIL_final_accuracy}. In class incremental learning, the final accuracy means the accuracy on all classes (100 classes for CIFAR-100 and 200 classes for Tiny ImageNet) after the training of the last phase. Applying ANCL improves the final accuracy of CL by 1-4 \%.  

\subsection{Hyperparameter Grid Search Result}\label{Appendix:F.4}

In this section, we present the grid search results of $\lambda$ and $\lambda_a$ on the benchmark (1)-(8) evaluated in \cref{table:mean_acc} and \cref{table:mean_acc_replay} in the main paper. For other hyperparameters required for specific methods, we follow the configuration of the original paper.  

\begin{table}[htbp]
\centering
\begin{tabular}{cccc}
\hline
Methods                                & Hyperparameter & (1) CIFAR-100/10 (10 tasks)                       & (2) CIFAR-100/20 (20 tasks)                      \\ \hline
EWC \cite{kirkpatrick2017overcoming}   & $\lambda$      & {[}5000, \textbf{10000}, 20000, 40000, 80000{]}   & {[}10000, 20000, 40000, \textbf{80000}, 160000{]} \\
w/ ANCL (ours)                         & $\lambda_a$    & {[}0.1, 1, \textbf{10}, 100, 1000, 10000{]}       & {[}0.1, 1, 10, \textbf{100}, 1000, 10000{]}\\\hline
MAS \cite{aljundi2018memory}           & $\lambda$      & {[}1, 5, 10, \textbf{50}, 100, 200{]}             & {[}1, 5, 10, 50, \textbf{100}, 200{]}\\
w/ ANCL (ours)                         & $\lambda_a$    & {[}0.1, 0.5, 1, \textbf{5}, 10, 50{]}             & {[}0.1, 0.5, \textbf{1}, 5, 10, 50{]} \\  \hline
LwF \cite{li2017learning}              & $\lambda$      & {[}0.1, 1, \textbf{10}, 100, 200, 400{]}          & {[}0.1, 1, \textbf{10}, 100, 200, 400{]} \\
w/ ANCL (ours)                         & $\lambda_a$    & {[}0.1, 0.5, \textbf{1}, 5, 10{]}                 & {[}0.1, 0.5, \textbf{1}, 5, 10{]}   \\  \hline
LFL \cite{jung2016less}                & $\lambda$      & {[}10, 100, 200, \textbf{400}, 800{]}             & {[}100, 200, 400, \textbf{800}, 1600{]} \\
w/ ANCL (ours)                         & $\lambda_a$    & {[}10, 50, \textbf{100}, 200, 400{]}              & {[}5, \textbf{10}, 50, 100, 200{]}  \\  \hline \hline
Methods                                & Hyperparameter & (3) TinyImagenet-200/10 (10 tasks)                & (4) TinyImagenet-200/20 (20 tasks)                      \\ \hline
EWC \cite{kirkpatrick2017overcoming}   & $\lambda$      & {[}5000, \textbf{10000}, 20000, 40000, 80000{]}   & {[}5000, \textbf{10000}, 20000, 40000, 80000{] } \\
w/ ANCL (ours)                         & $\lambda_a$    & {[}10, 100, 1000, \textbf{2000}, 3000{]}          & {[}10, 100, \textbf{1000}, 2000, 3000{]}\\\hline
MAS \cite{aljundi2018memory}           & $\lambda$      & {[}1, 5, \textbf{10}, 50, 100, 200{]}             & {[}1, 5, \textbf{10}, 50, 100, 200{]}\\
w/ ANCL (ours)                         & $\lambda_a$    & {[}0.05, \textbf{0.1}, 0.5, 1, 5, 10{]}           & {[}0.01, \textbf{0.05}, 0.1, 0.5, 1, 5, 10{]}  \\  \hline
LwF \cite{li2017learning}              & $\lambda$      & {[}0.1, 1, \textbf{10}, 100, 200, 400{]}          & {[}0.1, 1, \textbf{10}, 100, 200, 400{]}  \\
w/ ANCL (ours)                         & $\lambda_a$    & {[}0.1, 0.5, 1, \textbf{5}, 10{]}                 & {[}0.1, 0.5, 1, \textbf{5}, 10{]}    \\  \hline
LFL \cite{jung2016less}                & $\lambda$      & {[}100, 200, 400, \textbf{800}, 1600{]}           & {[}200, 400, 800, \textbf{1600}, 3200{]}\\
w/ ANCL (ours)                         & $\lambda_a$    & {[}10, 50, 100, \textbf{200}, 300{]}              & {[}10,  \textbf{50}, 100,200, 300{]}   \\  \hline
\end{tabular}
\caption{Hyperparameter search for CL and ANCL on benchmarks (1)-(4) in \cref{table:mean_acc} of the main paper. The set of parameters on which grid search is performed is shown, and selected hyperparameter is emphasized in bold.}
\label{table:gridsearch1}
\end{table}

\begin{table}[htbp]
\centering
\begin{tabular}{cccc}
\hline
Methods                                & Hyperparameter & (5) CIFAR-100/6 (1+5 phases)                      & (6) CIFAR-100/11 (1+10 phases)                       \\ \hline
iCaRL \cite{rebuffi2017icarl}          & $\lambda$      & {[}0.01, 0.05, 0.1, \textbf{0.5}, 1, 5{]}         & {[}0.01, 0.05, 0.1, 0.5, 1, \textbf{5}, 10{]} \\
w/ ANCL (ours)                         & $\lambda_a$    & {[}0.01, \textbf{0.05}, 0.1, 0.5, 1, 5{]}         & {[}0.01, 0.05, 0.1, \textbf{0.5}, 1, 5{]}\\\hline
BiC \cite{wu2019large}                 & $\lambda$      & {[}1, 5, \textbf{10}, 50, 100, 200{]}             & {[}1, 5, \textbf{10}, 50, 100, 200{]}\\
w/ ANCL (ours)                         & $\lambda_a$    & {[}0.1, 0.5, 1, \textbf{5}, 10, 50{]}             & {[}0.1, \textbf{0.5}, 1, 5, 10, 50{]} \\  \hline
LUCIR \cite{hou2019learning}           & $\lambda$      & {[}1, \textbf{5}, 10, 50, 100, 200{]}             & {[}1, \textbf{5}, 10, 50, 100, 200{]} \\
w/ ANCL (ours)                         & $\lambda_a$    & {[}0.1, 0.5, \textbf{1}, 5, 10{]}                 & {[}0.05 ,\textbf{0.1}, 0.5, 1, 5, 10{]}   \\  \hline
PODNet \cite{douillard2020podnet}      & $\lambda$      & {[}1, 5, \textbf{10}, 50, 100, 200{]}             & {[}1, 5, \textbf{10}, 50, 100, 200{]} \\
w/ ANCL (ours)                         & $\lambda_a$    & {[}0.1, 0.5, \textbf{1}, 5, 10{]}                 & {[}0.1, 0.5, \textbf{1}, 5, 10{]}  \\  \hline \hline
Methods                                & Hyperparameter & (7) TinyImagenet-200/11 (1+10 phases)             & (8) TinyImagenet-200/21 (1+20 phases)     \\ \hline
iCaRL \cite{rebuffi2017icarl}          & $\lambda$      & {[}0.01, 0.05, \textbf{0.1}, 0.5, 1, 5{]}         & {[}0.01, 0.05, 0.1, 0.5, \textbf{1}, 5{]} \\
w/ ANCL (ours)                         & $\lambda_a$    & {[}0.01, \textbf{0.05}, 0.1, 0.5, 1, 5{]}         & {[}0.01, 0.05, 0.1, \textbf{0.5}, 1, 5{]}\\\hline
BiC \cite{wu2019large}                 & $\lambda$      & {[}1, 5, \textbf{10}, 50, 100, 200{]}             & {[}1, 5, \textbf{10}, 50, 100, 200{]}\\
w/ ANCL (ours)                         & $\lambda_a$    & {[}0.05, \textbf{0.1}, 0.5, 1, 5, 10, 50{]}       & {[}0.005 , \textbf{0.01}, 0.05, 0.1, 0.5, 1, 5{]} \\  \hline
LUCIR \cite{hou2019learning}           & $\lambda$      & {[}1, 5, 10, \textbf{50}, 100, 200{]}             & {[}1, \textbf{5}, 10, 50, 100, 200{]} \\
w/ ANCL (ours)                         & $\lambda_a$    & {[}0.005, \textbf{0.01}, 0.05 ,0.1, 0.5, 1{]}     & {[}0.1, 0.5, 1, \textbf{5}, 10{]}   \\  \hline
PODNet \cite{douillard2020podnet}      & $\lambda$      & {[}1, 5, 10, 50, \textbf{100}, 200{]}             & {[}10, 50, 100, 200, \textbf{400}, 800{]} \\
w/ ANCL (ours)                         & $\lambda_a$    & {[}0.1, 0.5, \textbf{1}, 5, 10{]}                 & {[}0.1, 0.5, \textbf{1}, 5, 10{]}  \\  \hline 
\end{tabular}
\caption{Hyperparameter search for CL and ANCL on benchmarks (5)-(8) in \cref{table:mean_acc_replay} of the main paper.. The set of parameters on which grid search is performed is shown, and selected hyperparameter is emphasized in bold.}
\label{table:gridsearch2}
\end{table}

\section{Details on Stability-Plasticity Trade-off Analysis}\label{Appendix:G}

\subsection{Training Regime for Analysis}\label{Appendix:G.1}

\begin{figure}[htbp]
\centerline{\includegraphics[scale=.45]{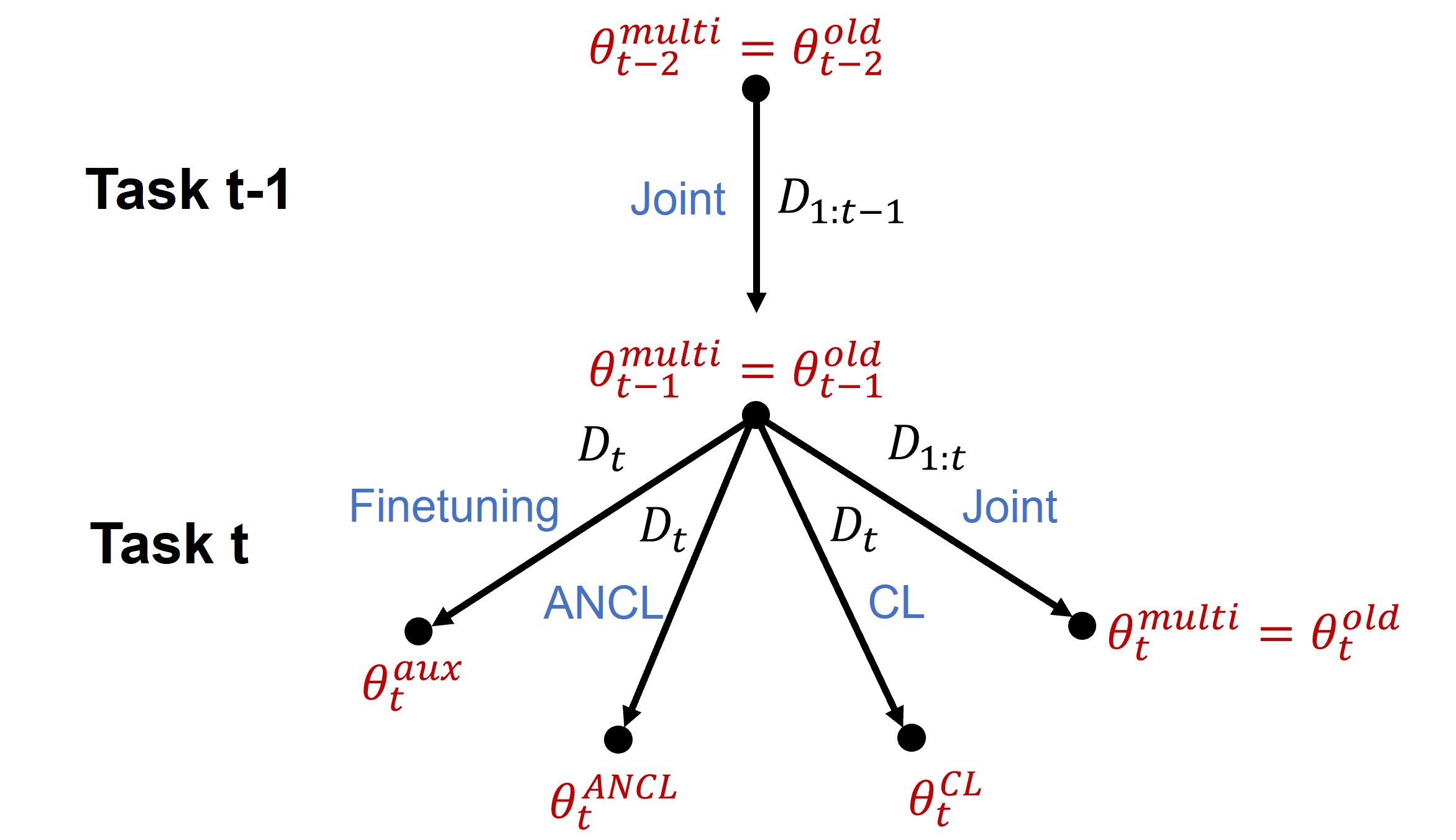}}
\caption{Training regime for analysis. $D_t$ stands for the dataset of task $t$ and $D_{1:t-1}$ means combined dataset from task $1$ to task $t-1$.}
\label{fig:training_regime}
\end{figure}

 In all analyses in the main text, the specific learning regime described in Fig.\ \ref{fig:training_regime} is adopted stemming from the regime from \cite{mirzadeh2020linear}. On task $t$, every model starts training from the multitask weights $\theta_{t-1}^{multi}$ which is trained on combined dataset $D_{1:t-1}$ ($i.e.$ train on $D_{1:t-1} = D_1 \cup \dots \cup D_{t-1}$). If we fine-tune the model on data $D_t$ without any regularization, it returns the auxiliary network $\theta_t^{aux}$ which will regularize ANCL later. ANCL and CL approach can be applied to obtain $\theta_t^{ANCL}$ and $\theta_t^{CL}$ each. Here, the multitask weights $\theta_{t-1}^{multi}$ on task $t-1$ works as the old network weights $\theta_{t-1}^{old}$ to regularize ANCL and CL. Finally, the initial weights are trained on data $D_{1:t}$ to train next multitask model $\theta_t^{multi}$ and it becomes next starting point for task $t+1$. We used fixed multitask weights as an initialization at the start of each task for fair comparison among all methods. Otherwise, every method will have different old and auxiliary network which end up confusing following analysis.

\subsection{Mode Connectivity Figures}\label{Appendix:G.2}

\begin{figure}[htbp]
\centerline{\includegraphics[scale=.45]{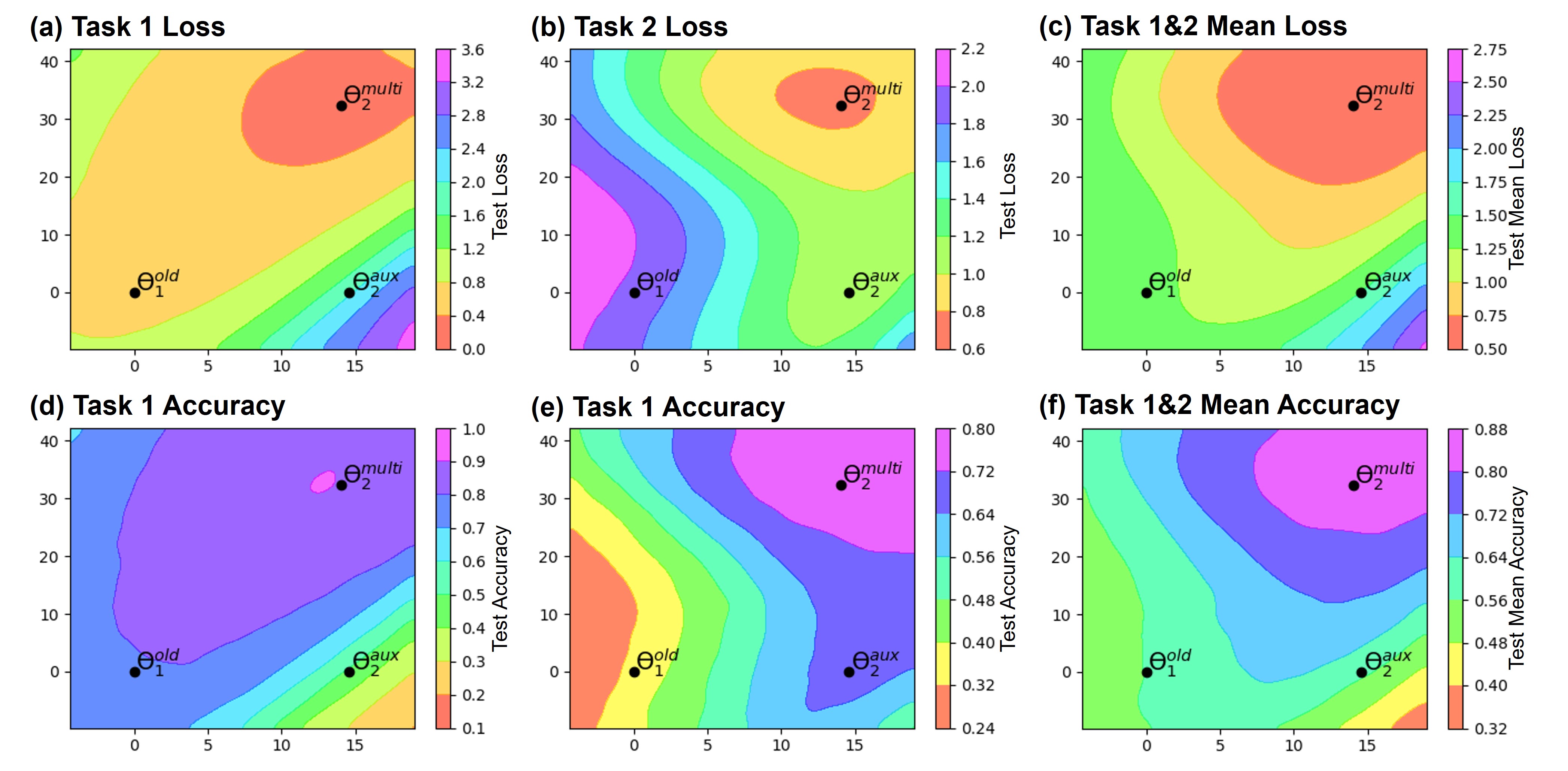}}
\caption{Test loss landscape (top low) and test accuracy landscape (bottom low) of task 1\&2 of benchmark (1) CIFAR-100/10. $\theta_1^{old}$, $\theta_2^{aux}$, and $\theta_2^{multi}$ are used to plot the two-dimensional subspace following \cite{mirzadeh2020linear}.}
\label{fig:mode_connectivity}
\end{figure}

Recent works \cite{draxler2018essentially, garipov2018loss} find a simple curve between the two local optima of deep neural networks (DNN) such that train loss and test error remain low along the curve. The simple linear path with low error can be visualized on the loss surface of DNN in gradient-based optimization setting. This path called \textit{Mode Connectivity} has been studied empirically and theoretically with some assumptions in different papers. 

Mode connectivity in continual learning is first investigated by \cite{mirzadeh2020linear}. They empirically show that mode connectivity holds between continual learning and multitask solutions when every training starts from same initialization. We also follow \cite{mirzadeh2020linear} to visualize mean accuracy landscape of the main paper and mode connectivity figures in \cref{fig:mode_connectivity} (details in \cref{Appendix:G.3}).

In \cref{fig:mode_connectivity}, $\theta_1^{old}$ is the old network weights trained on task $1$ only and $\theta_2^{aux}$ is the auxiliary network weights that is trained on task $2$ initialized by the old network. $\theta_2^{multi}$ is the multitask solution trained on the dataset of task $1$ and $2$. It is clear that the continual learning solutions ($\theta_1^{old}$ and $\theta_2^{aux}$) are linearly connected to the multitask solution ($\theta_2^{multi}$) by low loss and high accuracy path. Therefore, linear mode connectivity is valid in both loss and accuracy landscape. In the left and center column, the multitask weights are located in higher accuracy and lower loss contour compared to the continual learning weights. This is because the multitask model has a full access to previous datasets, consequently obtaining higher discriminating ability, while the continual models can learn only from the current dataset. The figures in right column visualize the mean loss and mean accuracy of task 1 and 2 and \cref{fig:mode_connectivity} (f) is used in mean accuracy landscape of the main paper. Higher accuracy and lower loss can be achieved in the middle of $\theta_1^{old}$ and $\theta_2^{aux}$. 

\subsection{The Visualization of Mean Accuracy Landscape}\label{Appendix:G.3}

In this Section, we explain how we visualize mean accuracy landscape in \cref{sec:trade-off_analysis} of main text and \cref{Appendix:G.2} following \cite{mirzadeh2020linear}. In order to build two basis vectors of the plane, we need three points $w_1$, $w_2$, and $w_3$ which corresponds to $\theta_1^{old}$, $\theta_2^{aux}$, and $\theta_2^{multi}$ respectively in mode connectivity figures. Each point refers to a high-dimensional weight vector which is obtained by flattening the weights of each layer in the neural network and then concatenating the flattened vectors including bias vector and batch normalization parameters. With three weight vectors at hand, we perform the following procedure:
\begin{enumerate}
    \item Calculate two basis vectors: $\Vec{u}=w_2-w_1$, and $\Vec{v}=w_3-w_1$.
    
    \item Orthogonalize the basis vectors by calculating
    \begin{align}
        \Vec{v} &= \Vec{v} - \norm{\Vec{v}}cos(\theta)\frac{\Vec{u}}{\norm{\Vec{u}}}\\
         &= \Vec{v}- \frac{\Vec{u} \cdot \Vec{v}}{\norm{\Vec{u}}^2}\Vec{u}
    \end{align}
    where $\theta$ denotes the angle between $\Vec{v}$ and $\Vec{u}$.
    
    \item Define a Cartesian coordinate system in $(x,y)$ plane that maps each coordinate to a parameter space by calculating $p(x,y)=w_1 + x\cdot \Vec{u} + y\cdot \Vec{v}$.
    
    \item For a defined grid on this coordinate system, we calculate the empirical loss and accuracy of each $(x,y)$ coordinate by reconstructing the weights of neural network from the high-dimensional vector $p(x,y)$. 
    
    \item The flattened weights of CL $w_{CL}$ can be expressed in the coordinate system $(x_{CL},y_{CL})$ by projecting the weight vector to $\Vec{u}$ and $\Vec{v}$ each:
    \begin{align}
        x_{CL} &= \frac{(w_{CL}-w_1)\cdot \Vec{u}}{\norm{\Vec{u}}^2},\\
        y_{CL} &= \frac{(w_{CL}-w_1)\cdot \Vec{v}}{\norm{\Vec{v}}^2}
    \end{align}
    In the same way, $(x_{ANCL},y_{ANCL})$ can be obtained from $w_{ANCL}$.
\end{enumerate}

The original weights $w_{CL}$ can be reconstructed from $p(x_{CL},y_{CL})$ and a residual vector $\Vec{R}$. The residual vector is aligned with the dimension orthogonal to both $\Vec{u}$ and $\Vec{v}$, thereby not being reflected in the coordinate system. Therefore, $w_{CL}$ can be fully expressed as below: 
\begin{align}
     w_{CL} = p(x_{CL},y_{CL}) + \Vec{R}=w_1 + x_{CL}\cdot \Vec{u} + y_{CL}\cdot \Vec{v} + \Vec{R}
\end{align}

\section{Mathematical Proofs}\label{Appendix:H}

\subsection{The Derivation of \cref{eq:aewc_kth_iter} in \cref{Appendix:E}} \label{appendix:iterative_formula_proof} 
In this section, we derive \cref{eq:aewc_kth_iter} by extending the derivation in \cite{lubana2021quadratic}. Let's start from rewriting the update rule of A-EWC from $k^{th}$ iteration to $(k+1)^{th}$ iteration with a learning rate $\eta$:
\begin{align}
    \theta_i^{(k+1)} = \theta_i^{(k)}  - \eta (\nabla_{\theta_i^{(k)}} \mathcal{L}_\text{t}(\theta^{(k)}) + \lambda F_{1:t-1,i} (\theta_i^{(k)}-\theta_{1:t-1,i}^{*}), + \lambda_a F_{t,i} (\theta_i^{(k)}-\theta_{t,i}^{*}))
    \label{appendix_eq:aewc_update}
\end{align}
where $\theta_i^{(k+1)}$ and $\theta_i^{(k)}$ are the $i^{th}$ model parameter on $k^{th}$ and $(k+1)^{th}$ iteration each such that $\theta_i\in \theta = (\theta_1,\dots,\theta_P)$ for the model weight $\theta\in\mathbb{R}^P$. $\nabla_{\theta_i^{(k)}} \mathcal{L}_\text{t}(\theta^{(k)})$ denotes the gradient of cross-entropy loss with respect to $\theta_i^{(k)}$ for a current task. $F_{1:t-1,i}$ and $F_{t,i}$ are the importance of EWC, the approximation of Fisher Information Matrix, for the old parameter $\theta_{1:t-1,i}^{*}$ and the auxiliary parameter $\theta_{t,i}^{*}$ respectively. Then, $\lambda$ and $\lambda_a$ adjust the strength of two regularizers. 

After rearranging \cref{appendix_eq:aewc_update}, the weight is updated as follows:
\begin{align}
    \theta_i^{(k+1)} &= (1-\alpha -\beta ) \theta_i^{(k)} + \alpha\theta_{1:t-1,i}^{*} + \beta \theta_{t,i}^{*} - \eta g_i^{(k)}  
    \label{appendix_eq:simplified_aewc_update}
\end{align}
where $\alpha = \eta\lambda F_{1:t-1,i}$, $\beta = \eta\lambda_a F_{t,i}$ and $g_i^{(k)} = \nabla_{\theta_i^{(k)}} \mathcal{L}_\text{t}(\theta^{(k)})$ are applied. From above update rule, we will prove that \cref{eq:aewc_kth_iter} holds true for all $k^{th}$ iteration through mathematical induction. At 1st iteration, we have following equation by directly substituting $k=0$ in \cref{appendix_eq:simplified_aewc_update}:
\begin{align}
    \theta_i^{(1)} &= (1-\alpha -\beta ) \theta_{1:t-1,i}^{*} + (\alpha\theta_{1:t-1,i}^{*} + \beta \theta_{t,i}^{*}) - \eta g_i^{(0)}  
\end{align}
where the model for $t^{th}$ task is initialized by the final optimal weight of $(t-1)^{th}$ task such that $\theta_i^{(0)} = \theta_{1:t-1,i}^{*}$. Then, it automatically satisfies \cref{eq:aewc_kth_iter} as following:
\begin{equation}
    \theta_i^{(1)} = (1-\alpha-\beta) \theta_{1:t-1,i}^{*} +\sum_{l=0}^{0}(1-\alpha-\beta)^l (\alpha\theta_{1:t-1,i}^{*} + \beta\theta_{t,i}^{*} ) - \sum_{l=0}^{0} [(1-\alpha-\beta)^{(1-l-1)}\eta] g_i^{(l)}
\end{equation}
Assume \cref{eq:aewc_kth_iter} holds true at $k^{th}$ iteration:
\begin{align}
    \theta_i^{(k)} = (1-\alpha-\beta)^k \theta_{1:t-1,i}^{*} +\sum_{l=0}^{k-1}(1-\alpha-\beta)^l (\alpha\theta_{1:t-1,i}^{*} + \beta\theta_{t,i}^{*} ) - \sum_{l=0}^{k-1} [(1-\alpha-\beta)^{(k-l-1)}\eta] g_i^{(l)}.
\end{align}
Then, at $(k+1)^{th}$ iteration, we have:
\begin{equation}
    \theta_i^{(k+1)} = (1-\alpha -\beta ) \theta_i^{(k)} + \alpha\theta_{1:t-1,i}^{*} + \beta \theta_{t,i}^{*} - \eta g_i^{(k)}
\end{equation}
\begin{multline}
  = (1-\alpha -\beta ) ((1-\alpha-\beta)^k \theta_{1:t-1,i}^{*} +\sum_{l=0}^{k-1}(1-\alpha-\beta)^l (\alpha\theta_{1:t-1,i}^{*} + \beta\theta_{t,i}^{*} )  \\ - \sum_{l=0}^{k-1} [(1-\alpha-\beta)^{(k-l-1)}\eta] g_i^{(l)}) + \alpha\theta_{1:t-1,i}^{*} + \beta \theta_{t,i}^{*} - \eta g_i^{(k)}   
\end{multline}
\begin{multline}
  = (1-\alpha-\beta)^{(k+1)} \theta_{1:t-1,i}^{*} +\sum_{l=0}^{k-1}(1-\alpha-\beta)^{(l+1)} (\alpha\theta_{1:t-1,i}^{*} + \beta\theta_{t,i}^{*} )  \\ - \sum_{l=0}^{k-1} [(1-\alpha-\beta)^{(k+1-l-1)}\eta] g_i^{(l)} + \alpha\theta_{1:t-1,i}^{*} + \beta \theta_{t,i}^{*} - \eta g_i^{(k)}   
\end{multline}
\begin{equation}
    = (1-\alpha-\beta)^{(k+1)} \theta_{1:t-1,i}^{*} +\sum_{l=0}^{k}(1-\alpha-\beta)^l (\alpha\theta_{1:t-1,i}^{*} + \beta\theta_{t,i}^{*} ) - \sum_{l=0}^{k} [(1-\alpha-\beta)^{(k+1-l-1)}\eta] g_i^{(l)}.
\end{equation}
The last equality satisfies \cref{eq:aewc_kth_iter} at $(k+1)^{th}$ iteration which ends our proof.

\subsection{The Derivation of \cref{eq:hinton_KD} in \cref{Appendix:E}} \label{appendix:hinton_proof} 

For the completeness of our paper, we expand the derivation of \cref{eq:hinton_KD} according to \cite{hinton2015distilling}. We first rewrite the loss of LwF in \cref{eq:lwf_loss} and define the second term as distillation loss $\mathcal{L}_\text{D}$:
\begin{align}
    \mathcal{L}_\text{LwF} &= \mathcal{L}_\text{t}(\theta) + \lambda \sum_{c=1}^{C_{1:t}} -y^c(x_j;\theta_{1:t-1}^*) \log{y^c(x_j;\theta)},\\
   \mathcal{L}_\text{D} &= \sum_{c=1}^{C_{1:t}} -y^c(x_j;\theta_{1:t-1}^*) \log{y^c(x_j;\theta)}.
\end{align}
Then, we take a derivative on the distillation loss $\mathcal{L}_\text{D}$ with respect to the logit $\boldsymbol{o}^h(x_j;\theta)$: 
\begin{align}
    \nabla_{\boldsymbol{o}^h(x_j;\theta)} \mathcal{L}_\text{D} &=  \frac{\partial}{\partial \boldsymbol{o}^h(x_j;\theta)} \sum_{c=1}^{C_{1:t}} -y^c(x_j;\theta_{1:t-1}^*) \log{y^c(x_j;\theta)}\\
    &= \sum_{c=1}^{C_{1:t}} -y^c(x_j;\theta_{1:t-1}^*) \frac{\partial \log{y^c(x_j;\theta)}}{\partial \boldsymbol{o}^h(x_j;\theta)} \\
    &= \sum_{c=1}^{C_{1:t}} -  \frac{y^c(x_j;\theta_{1:t-1}^*)}{y^c(x_j;\theta)} \frac{\partial y^c(x_j;\theta)}{\partial \boldsymbol{o}^h(x_j;\theta)} \\
    &= - \frac{y^h(x_j;\theta_{1:t-1}^*)}{y^h(x_j;\theta)} \frac{\partial y^h(x_j;\theta)}{\partial \boldsymbol{o}^h(x_j;\theta)} + \sum_{c \ne h} -\frac{y^c(x_j;\theta_{1:t-1}^*)}{y^c(x_j;\theta)} \frac{\partial y^c(x_j;\theta)}{\partial \boldsymbol{o}^h(x_j;\theta)}\\
    &= \frac{1}{\tau} ( y^h(x_j;\theta) -  y^h(x_j;\theta_{1:t-1}^*)) \label{eq:temp_KD_der}
\end{align}
where in the last equality we applied the following:
\begin{align}
    \frac{\partial y^c(x_j;\theta)}{\partial \boldsymbol{o}^h(x_j;\theta)} = \partial( \frac{e^{\boldsymbol{o}^c(x_j;\theta)/\tau}}{\sum_k e^{\boldsymbol{o}^k(x_j;\theta)/\tau}})/ \partial \boldsymbol{o}^h(x_j;\theta) = 
\begin{cases}
\frac{1}{\tau} y^h(x_j;\theta)(1-y^h(x_j;\theta)) , &\text{ if } c=h\\
-\frac{1}{\tau} y^c(x_j;\theta)y^h(x_j;\theta),  &\text{ otherwise }
\end{cases}
\end{align}
If we take derivative on the loss of LwF with respect to $\theta$:
\begin{align}
    \nabla_{\theta} \mathcal{L}_\text{LwF} &= \nabla_{\theta} \mathcal{L}_\text{t}(\theta) +  \lambda \frac{\partial}{\partial \theta} \mathcal{L}_\text{D} \\
    &= \nabla_{\theta} \mathcal{L}_\text{CE}(\theta) +  \lambda  \frac{\partial\boldsymbol{o}(x_j;\theta)}{\partial \theta} \frac{\partial \mathcal{L}_\text{D}}{\partial \boldsymbol{o}(x_j;\theta)}\\
&= \nabla_{\theta} \mathcal{L}_\text{CE}(\theta) + \frac{\lambda}{\tau}  (y(x_j;\theta) -  y(x_j;\theta_{1:t-1}^*)) \nabla_{\theta} \boldsymbol{o}(x_j;\theta)
\end{align}
In the last equality, we applied \cref{eq:temp_KD_der} and it ends our derivation of \cref{eq:hinton_KD}.

\end{document}